\documentclass[acmsmall]{acmart}
\usepackage{multirow}
\usepackage{makecell}
\usepackage{booktabs}
\usepackage{soul}

\soulregister{\cite}7
\soulregister{\ref}7
\AtBeginDocument{%
  \providecommand\BibTeX{{%
    \normalfont B\kern-0.5em{\scshape i\kern-0.25em b}\kern-0.8em\TeX}}}

\setcopyright{acmcopyright}
\copyrightyear{2022}
\acmYear{2022}
\acmDOI{XXXXXXX.XXXXXXX}

\acmJournal{TIST}
\acmVolume{XX}
\acmNumber{X}
\acmArticle{111}
\acmMonth{12}

\begin{document}

\title{Recent Few-Shot Object Detection Algorithms: A Survey with Performance Comparison}

\author{Tianying Liu}
\email{liutianying@tongji.edu.cn}
\affiliation{%
  \institution{Tongji University}
  \department{Department of Computer Science and Technology}
  \postcode{201804}
  \department{Project Management Office of China National Scientific Seafloor Observatory}
  \postcode{200092}
  \city{Shanghai}
  \country{China}
}

\author{Lu Zhang}
\email{l\_zhang19@fudan.edu.cn}
\affiliation{%
  \institution{Fudan University}
  \department{Shanghai Key Lab of Intelligent Information Processing}
  \department{School of Computer Science}
  \postcode{200438}
  \city{Shanghai}
  \country{China}
}

\author{Yang Wang}
\email{tongji\_wangyang@tongji.edu.cn}
\affiliation{%
  \institution{Tongji University}
  \department{Department of Computer Science and Technology}
  \postcode{201804}
  \department{Project Management Office of China National Scientific Seafloor Observatory}
  \postcode{200092}
  \city{Shanghai}
  \country{China}
}

\author{Jihong Guan}
\email{jhguan@tongji.edu.cn}
\affiliation{%
  \institution{Tongji University}
  \department{Department of Computer Science and Technology}
  \postcode{201804}
  \department{Project Management Office of China National Scientific Seafloor Observatory}
  \postcode{200092}
  \city{Shanghai}
  \country{China}
}

\author{Yanwei Fu}
\email{yanweifu@fudan.edu.cn}
\affiliation{%
  \institution{Fudan University}
  \department{Shanghai Key Lab of Intelligent Information Processing}
  \department{School of Data Science}
  \postcode{200433}
  \city{Shanghai}
  \country{China}
}

\author{Jiajia Zhao}
\email{zhaojiajia1982@gmail.com}
\affiliation{%
  \institution{Science and Technology on Complex System Control and Intelligent Agent Cooperation Laboratory}
  \postcode{100074}
  \city{Beijing}
  \country{China}
}

\author{Shuigeng Zhou}
\authornote{Corresponding Author}
\email{sgzhou@fudan.edu.cn}
\affiliation{%
  \institution{Fudan University}
  \department{Shanghai Key Lab of Intelligent Information Processing}
  \department{School of Computer Science}
  \postcode{200438}
  \city{Shanghai}
  \country{China}
}

\authorsaddresses{%
Authors' addresses: T.~Liu, Y.~Wang and J.~Guan, Department of Computer Science and Technology,
Tongji University, Shanghai 201804, China;
and the Project Management Office of China National Scientific Seafloor Observatory,
Tongji University, Shanghai 200092, China;
emails: \{liutianying, tongji\_wangyang, jhguan\}@tongji.edu.cn;
L.~Zhang and S.~Zhou, Shanghai Key Lab of Intelligent Information Processing,
and the School of Computer Science,
Fudan University, Shanghai 200438, China;
emails: e-mail: \{l\_zhang19, sgzhou\}@fudan.edu.cn;
Y.~Fu, Shanghai Key Lab of Intelligent Information Processing,
and the School of Data Science,
Fudan University, Shanghai 200433, China;
email: yanweifu@fudan.edu.cn.
J. Zhao, Science and Technology on Complex System Control and Intelligent Agent Cooperation Laboratory, Beijing 100074, China;
email: zhaojiajia1982@gmail.com
}

\renewcommand{\shortauthors}{Liu et al.}

\begin{abstract}
  The generic object detection~(GOD) task has been successfully tackled  by recent deep neural networks,
  trained by an avalanche of annotated training samples from some common classes.
  However, it is still non-trivial to generalize these object detectors to the novel long-tailed object classes,
  which have only few labeled training samples. To this end, the Few-Shot Object Detection (FSOD) has been topical recently,
  as it mimics the humans' ability of \textit{learning to learn}, and intelligently transfers the learned generic object knowledge from the common heavy-tailed,
  to the novel long-tailed object classes.
  Especially, the research in this emerging field has been flourishing in recent years with various benchmarks, backbones, and methodologies proposed.
  To review these FSOD works, there are several insightful FSOD survey articles~\cite{huangqi2021survey, leng2021comparative, kohler2021few, huang2021survey} that
  systematically study and compare them as the groups of fine-tuning/transfer learning, and meta-learning methods.
  In contrast, we review the existing FSOD algorithms from a new perspective under a new taxonomy based on their contributions,
  i.e., data-oriented, model-oriented, and algorithm-oriented.
  Thus, a comprehensive survey with performance comparison is conducted on recent achievements of FSOD.
  Furthermore, we also analyze the technical challenges, the merits and demerits of these methods, and envision the future directions of FSOD.
  Specifically, we give an overview of FSOD, including the problem definition, common datasets, and evaluation protocols.
  The taxonomy is then proposed that groups FSOD methods into three types.
  Following this taxonomy, we provide a systematic review of the advances in FSOD.
  Finally, further discussions on performance, challenges, and future directions are presented.

\end{abstract}

\ccsdesc[500]{Computing methodologies~Object detection}
\ccsdesc[300]{Computing methodologies~Neural networks}
\ccsdesc[500]{Computing methodologies~Transfer learning}

\keywords{few-shot learning, meta-learning, survey}

\maketitle

\section{Introduction}
Object detection has made great progress in recent years by leveraging deep neural networks (e.g. deep convolutional neural networks (CNNs))~\cite{simonyan2014very, he2016deep} trained on massive labeled training samples.
Typically, these detectors have to rely on an avalanche of annotated examples, and are expected to have significant performance dropping on the novel long-tailed object classes, which usually only have few labeled samples.
In particular, the frequency of object occurrence in the wild follows a long-tail distribution~\cite{longtailed2011}: the common heavy-tailed objects appear quite often,
while much more novel objects such as extinct species are long-tailed distributed in sense that they are either labor-intensive or expensive to be annotated.

\begin{figure}[!t]
	\centering
	\includegraphics[width=0.9\textwidth]{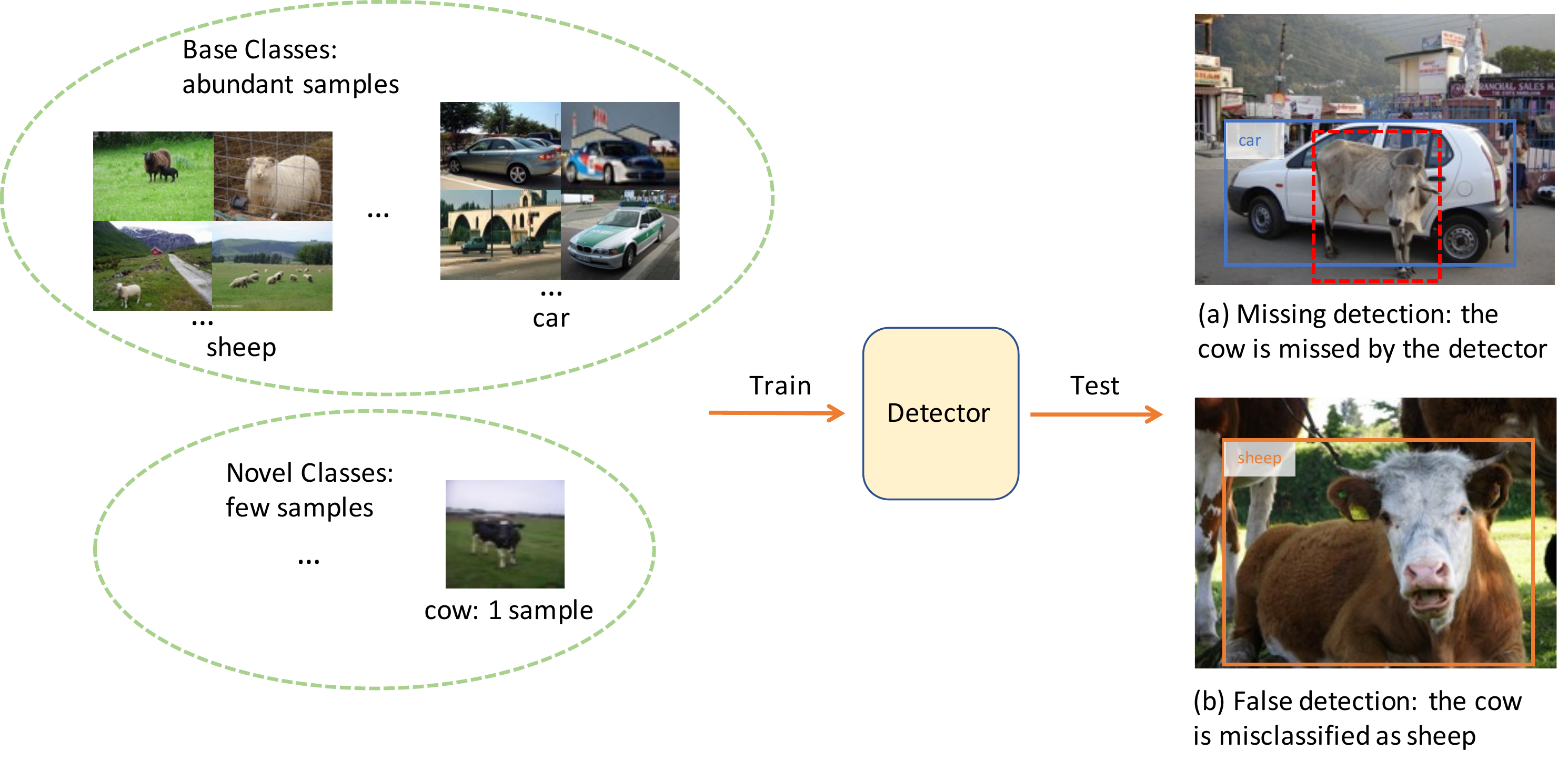}
	\caption{Examples of missing/false detection under the 1-shot setting. The detector is trained on both base classes and novel classes. Each base class contains abundant annotated samples, while each novel class has only a training instance. Due to the lack of sufficient training samples, the model always performs poorly on novel classes. (a) Novel objects may be missed during inference. Here, the red dashed box indicates the missing cow. (b) Novel objects may be misclassified as base classes. Here, the cow is wrongly categorized into sheep).}
	\label{fig:example}
\end{figure}

To mimic the humans' ability of easily \textit{learning to learn} novel concepts, the research area of  Few-Shot Learning (FSL)~\cite{fe2003bayesian} has received increasing interest.
FSL targets at learning from a limited number of training samples with supervised information.
As in~\cite{koch2015siamese, vinyals2016matching, bertinetto2016learning, santoro2016meta, snell2017prototypical, satorras2018few, ravi2016optimization, gidaris2018dynamic, sung2018learning},
most of the previous FSL works focuses on the  Few-Shot Classification (FSC) task, i.e., recognizing classes with limited samples in the query images, and to a lesser extent Few-Shot Object Detection (FSOD)~\cite{chenLSTD2018}.
FSOD aims at intelligently transferring previously learned knowledge on common heavy-tailed objects or the generic object detectors to help future object detection tasks on the novel long-tailed object categories.
In particular, the major challenge of FSOD comes from building a detector capable of detecting novel objects with only a few labeled examples.
The model is prone to overfit without sufficient training samples. It is hard to estimate the true data distribution with a few samples. Therefore, novel objects with various appearances are easy to be missed by the detector.
Moreover, highly alike classes always confuse the detectors owing to the lack of sufficient training samples to obtain the distinguishable features~\cite{liMaxMargin2021, liFewShot2021a}. The novel objects are prone to be wrongly classified into some alike base categories.
Fig.~\ref{fig:example} illustrates two failed detection examples of a typical FSOD method TFA~\cite{wangFrustratingly2020} under the 1-shot setting, where each novel class has only one training sample~\cite{wangFrustratingly2020}.
Fig.~\ref{fig:example}(a) shows that the query cow is missed by the detector as it appears very different from the support cow and the other support classes.
Fig.~\ref{fig:example}(b) indicates that when the query cow looks like a base class sheep, it is very likely to be misclassified as that.

The common practices of existing methods are to solve the FSOD problem as the FSC problem.
Thus, many recent FSOD survey articles~\cite{huangqi2021survey, leng2021comparative, kohler2021few, huang2021survey} give insightful discussions on FSOD works by grouping them into fine-tuning/transfer learning, and meta-learning methods.
Particularly, as an important learning paradigm in FSC, meta-learning has been also widely adopted to learn the meta-detectors in many FSOD methods,
such as CoAE~\cite{hsiehOneShot2019a}, RepMet~\cite{karlinskyRepMet2019}, MetaDet~\cite{wangMetaLearning2019}, Meta R-CNN~\cite{yanMeta2019}, and FSRW~\cite{kangFewShot2019}, etc.
Meta detectors are designed to learn the category-agnostic knowledge via feature re-weighting or specific metrics, generally trained with the \textit{episodes} sampled from base classes to explicitly simulate the few-shot scenario.
On the other hand, the alternative approach to FSOD is the two-stage fine-tuning-based detector, which first pre-trains the model on base classes, and then fine-tunes this model on the support set with some parameters frozen~\cite{wangFrustratingly2020}.
This simple method shows the potential in improving the FSOD performance~\cite{sunFSCE2021, fanGeneralized2021, wuUniversalPrototype2021}.
To alleviate the impact of limited training data, some FSOD approaches utilize extra datasets to import prior knowledge~\cite{zhuSemantic2021}, while others propose special data augmentation modules to generate more training samples~\cite{liTransformation2021, zhangHallucination2021}.

Although significant progress has been made in FSOD, there is still a far cry from FSOD to the human visual system.
First, the majority of FSOD methods are based on the generic object detection~(GOD) frameworks, such as Faster R-CNN~\cite{ren2015faster} and YOLO~\cite{redmon2016you},
which rely on high-quality region proposals to obtain satisfactory detection results. However, in the few-shot scenario, it is hard to generate such proposals for novel classes with a few training samples~\cite{fanFewShot2020, zhangMetaDETR2021}, which limits the performance upper-bound of FSOD.
Second, detectors are inclined to consider irrelevant features due to inadequate support samples, leading to higher intra-class variations.
As visual appearances such as colors and textures vary in different instances of the same category, it is more challenging for few-shot detectors to classify objects in novel classes~\cite{huDense2021, liMaxMargin2021, wuUniversalPrototype2021}.
Therefore, there is still a long way to pursue the FSOD task.

Differently from previous survey works, we review these FSOD algorithms from a new perspective and with a new taxonomy based on their contributions, i.e., data-oriented, model-oriented, and algorithm-oriented.
Thus, a comprehensive survey with performance comparison is conducted on the recent achievements of FSOD. Furthermore,
we further analyze the technical challenges, the merits and demerits of these methods, and envision the future directions of FSOD. Specifically,
we review the recent advances in FSOD and discuss the potential research directions.
The remainder of this paper is organized as follows.
Section~\ref{sec:related-reviews} gives a brief introduction to the related surveys.
Section~\ref{sec:preliminaries} introduces the preliminaries, mainly FSL and generic object detection.
Section~\ref{sec:overview} presents an overview of FSOD, including problem definition, common datasets, metrics, and our taxonomy of FSOD methods.
Following the proposed taxonomy, Section~\ref{sec:fsod} presents a detailed review of each kind of existing FSOD method.
Finally, Section~\ref{sec:analysis} conducts a performance comparison of the existing FSOD methods, while Section~\ref{sec:challenges} and Section~\ref{sec:directions} highlight main challenges and future directions, respectively.

\section{Related Reviews}\label{sec:related-reviews}
\subsection{Reviews on Few-Shot Learning~(FSL)}
In recent years, FSL has been an active topic and there are already several reviews~\cite{wang2020generalizing, bendre2020learning, kadam2018review, jadon2020overview, li2021concise, parnami2022learning}.
Wang et al.~\cite{wang2020generalizing} discussed the learning issues in FSL and pointed out that the core issue is the unreliable empirical risk minimizer.
They categorized FSL methods from three perspectives, i.e., data, model, and algorithm,
which is a reasonable way to understand how prior knowledge is leveraged to improve few-shot classification performance.
Jadon~\cite{jadon2020overview} reviewed some widely-used deep learning architectures for FSL and mentioned some relevant fields: semi-supervised, imbalanced, and transfer learning.
But this survey described only the technical details of the related methods without deep discussions about FSL.
Kadam et al.~\cite{kadam2018review} highlighted several challenges of FSL: optimization and regularization of large networks, strong generalization, learning multitasking, and theoretical bounds for FSL.
The methods were grouped into the data-bound and learning-bound approaches, and the corresponding strategies were summarized. But the specific methods were not given.
Bendre et al.~\cite{bendre2020learning} and Li et al.~\cite{li2021concise} analyzed FSL methods from the perspective of meta-learning.
Bendre et al.~\cite{bendre2020learning} combined few-shot, one-shot, and zero-shot techniques and group them into data-augmentation-based, embedding-based, optimization-based, and semantics-based methods.
For each type, several models were elaborated, and their performances were compared.
Li et al.~\cite{li2021concise} summarized few-shot meta-learning approaches as learning an initialization, generation of parameters, learning an optimizer, and memory-based methods.
Although meta-learning is a significant part of FSL, other techniques such as transfer learning are also needed to be reviewed to make the survey more comprehensive.
Parnami et al.~\cite{parnami2022learning} clustered FSL techniques into meta-learning-based and non-meta-learning-based methods. Besides describing each model detailedly, the advantages and disadvantages of each type were concluded. They also summarized the progress of , but presented performance only under a certain setting of 5-way 1-shot image classification task.\\
\indent In addition to reviews on few-shot image classification, there are also reviews focusing on other few-shot scenarios.
Yin~\cite{yin2020meta} focused on Natural Language Processing (NLP) and introduced several meta-learning-based methods, which were categorized into two classes: meta-learning on different domains of the same problem and meta-learning on diverse problems.
Zhang et al.~\cite{zhang2022few} reviewed the progress of few-shot learning on graphs, and summarized the methods according to three problems: node-level FSL, edge-level FSL, and graph-level FSL. For each problem, there are two types of methods, i.e., metric-based and optimization-based.\\
\indent The surveys above all focus on classification tasks. In this paper, we concentrate on the applications of FSL to the object detection problem.

\subsection{Reviews on Generic Object Detection~(GOD)}
As one of the most fundamental problems of computer vision, there is a variety of object detection approaches proposed every year, and various surveys have been published~\cite{liu2020deep, jiao2019survey, zhao2019object, sultana2020review, zou2019object, oksuz2020imbalance, padilla2020survey, tong2020recent}. Among them,
Liu et al.~\cite{liu2020deep}, Jiao et al.~\cite{jiao2019survey}, Zhao et al.~\cite{zhao2019object}, and Sultana et al.~\cite{sultana2020review} provided a comprehensive analysis on deep learning for object detection, including milestone object detectors, popular datasets, metrics, several subfields or applications of object detection, etc.
Zou et al.~\cite{zou2019object} extensively reviewed the technical evolution of object detection in the past 20 years, including both traditional object detection methods and deep learning-based detection methods.
They also paid attention to detection speed-up techniques and challenges.
Oksuz et al.~\cite{oksuz2020imbalance} surveyed object detection from the perspective of imbalance problems, and grouped the problems into four main types: class imbalance, scale imbalance, objective imbalance, and bounding-box imbalance.
Padilla et al.~\cite{padilla2020survey} compared the most widely used metrics for performance evaluation, and elaborated on their differences, applications, and main concepts.
Tong et al.~\cite{tong2020recent} comprehensively reviewed the existing small object detectors, which is one of the most challenging problems of object detection.
Although the aforementioned surveys analyze the recent advances in object detection from different perspectives, they do not cover the achievements of FSOD.

\subsection{Reviews on Few-Shot Object Detection}
Previous reviews commonly categorize FSOD methods into fine-tuning-based (or transfer learning-based) and meta-learning-based methods~\cite{antonelli2021few, huangqi2021survey, leng2021comparative, kohler2021few, huang2021survey}.
\begin{itemize}
	\item \textbf{Fine-tuning/Transfer learning}.
	Fine-tuning-based methods generally pre-train the model on the \textit{base set}, which contains abundant labeled data, and then fine-tune the model on the \textit{support set}, which only contains a few labeled data for each class.
	The pre-trained parameters can provide the model with a good initialization to obtain acceptable performance on the \textit{support set}.
	A typical example is TFA~\cite{wangFrustratingly2020}, which only fine-tunes several layers while keeping other layers frozen when training on the \textit{support set}.
	It demonstrates the potential of fine-tuning-based methods.
	\item \textbf{Meta-learning}.
	Meta-learning-based methods typically subsample the \textit{base set} as \textit{episodes} to simulate the few-shot scenario.
	Each \textit{episode} contains the same number of samples for each class as the \textit{support set}.
	After training on the \textit{episodes}, the model obtains category-agnostic notions, which are helpful to detect objects of novel classes.
	Feature re-weighting~\cite{kangFewShot2019, wangMetaLearning2019}, metric-learning~\cite{karlinskyRepMet2019, yangRestoring2020}, prototype matching~\cite{wuMetaRCNN2020, han2021meta} networks are commonly-used approaches to improve the performance of meta-detectors.
\end{itemize}

Based on the above two categories, Antonelli~\cite{antonelli2021few} added two other types, i.e., data augmentation and distance metric learning, to fully describe which aspect the FSOD methods are improved by leveraging prior knowledge.
FSOD methods are also grouped into self-supervised, semi-supervised, weakly-supervised, and limited-supervised methods based on whether or how much the supervisory signal is used~\cite{huangqi2021survey, leng2021comparative, huang2021survey}.

However, existing classification taxonomies above are inadequate to discriminate the broad variety of FSOD methods.
For example, fine-tuning-based and meta-learning-based methods may have similar ideas in model design.
Therefore, 
in this survey, we try to introduce a new taxonomy of FSOD methods according to their major contributions from three perspectives: data-oriented, model-oriented, and algorithm-oriented, which is a clearer way to understand the motivation of each method and helpful for envisioning the future directions.
Furthermore, we analyze and compare the performances of major existing FSOD methods on commonly-used datasets.

\section{Preliminaries}\label{sec:preliminaries}
In this section, we briefly introduce FSL and generic object detection, including the problem definition, basic concepts, and major recent advances.

\subsection{Few-Shot Learning (FSL)}
FSL is a sub-area of machine learning, aiming at learning from a few annotated samples.
Traditional supervised learning methods often encounter significant performance drop when the supervised information of a specific task is very limited.
Therefore, FSL approaches always utilize the prior knowledge from other available sources to make the learning of the target feasible~\cite{wang2020generalizing}.
A typical application of FSL is Few-Shot image Classification (FSC).
Most commonly, the \textit{N}-way \textit{K}-shot problem is considered, where there are \textit{N} novel (or unseen) classes with \textit{K} labeled images for each class.
The \textit{N}$\times$\textit{K} images constitute the \textit{support set}.
And the goal is to recognize the query images belonging to these classes~\cite{vinyals2016matching}.
Especially, when \textit{K} is limited to 1, FSL becomes \textit{One-Shot Learning}.
And when \textit{K} is set to 0, FSL turns to \textit{Zero-Shot Learning}, which requires information from other modalities such as word embeddings.
Roughly, FSL methods can be categorized into three types, i.e., transfer learning, meta-learning, and data augmentation methods.

The core idea of transfer learning is to apply the knowledge learned from a certain source domain to a different but related task, which is easier than learning from scratch~\cite{thrun1996learning}.
Dong et al.~\cite{dong2018domain} proposed a domain adaption framework in one-shot learning.
Chen et al.~\cite{chen2019closer} showed that a baseline with standard fine-tuning is effective to solve the FSL problem.
Medina et al.~\cite{medina2020self} presented a self-supervised prototypical transfer learning approach, which performs well even when domain shift exists.
Transmatch~\cite{yu2020transmatch} developed a novel transfer learning framework for semi-supervised FSL, utilizing the auxiliary information of both base classes and novel classes.

Meta-learning methods are designed to rapidly generalize to new tasks that have never been seen during training.
Generally, the training process is the same way as testing.
Given a training set with abundant labeled data, meta-learning methods firstly sample \textit{episodes} from the training set to mimic the \textit{N}-way \textit{K}-shot scenario.
After meta-training on the sampled \textit{episodes}, the model is able to obtain the task-agnostic meta-level knowledge, which can be adapted to novel classes to tackle the target problem.
There are three types of meta-learning methods, i.e., metric-based, model-based, and optimization-based ones.
\begin{itemize}
\item \textbf{Metric-based} approaches aim to learn a kernel function that measures the similarity between two data samples~\cite{koch2015siamese, vinyals2016matching, snell2017prototypical, sung2018learning}.
For instance, Prototypical Networks~\cite{snell2017prototypical} learn a kernel function to project each input into a new feature space, and the class of the query image is determined by its distance to the support prototypes.
\item \textbf{Model-based} methods concentrate on designing models capable of updating their parameters with only a few samples.
Some models import external memory module such as Neural Turing Machine (NTM)~\cite{graves2014neural} and Memory Networks~\cite{weston2014memory} to store the information, which is helpful for fast learning~\cite{santoro2016meta, cai2018memory, ramalho2019adaptive}.
For instance, Santoro et al.~\cite{santoro2016meta} proposed an NTM-based framework to memorize the information longer.
Meta Networks~\cite{munkhdalai2017meta} utilize a neural network to predict the parameters of another network to accelerate the training process.
Gidaris et al.~\cite{gidaris2018dynamic} presented a weight generator for novel categories to enhance the original classifier.
\item \textbf{Optimization-based} algorithms intend to make the gradient-based optimization process adaptable to the few-shot scenario.
Ravi et al.~\cite{ravi2016optimization} modeled the meta-learner as Long Short-Term Memory (LSTM)~\cite{hochreiter1997long} to mimic the gradient-based update.
MAML~\cite{finn2017model} is another representative model capable of initializing the network with good parameters, which can be plugged into any model learning through gradient descent.
Reptile~\cite{nichol2018first} is another well-known optimization algorithm similar to MAML.
\end{itemize}

FSL methods based on data augmentation try to enrich the supervised information to mitigate the overfitting problem and obtain more reliable models.
Traditional data augmentation approaches with handcrafted rules are widely adopted as pre-processing, such as translation, flipping, cropping, scaling, and rotation.
However, these simple operations are not enough to capture all kinds of invariants.
Thus, some advanced augmentation methods are proposed.
One set of methods focuses on the feature space. Hariharan et al.~\cite{hariharan2017low} hallucinated the feature vectors for the training set to generate additional training examples for data-starved classes.
Zhang et al.~\cite{zhang2019few} utilized a saliency map to acquire foreground and background information of the input image and then generated hallucinated data points in the feature space.
AFHN~\cite{li2020adversarial} hallucinates diverse and discriminative features based on a Generative Adversarial Networks (GAN)~\cite{goodfellow2014generative}.
Other methods intend to produce additional samples.
Wang et al.~\cite{wang2018low} used a GAN to generate an augmented training set, which is jointly trained with the meta-learner.
Chen et al.~\cite{chen2019image} added an image deformation sub-network to generate extra samples.
DMAS~\cite{gui2021learning} imports a mentor model on base classes for directing the hallucinator to produce high-quality samples.

\subsection{Generic Object Detection (GOD)}
Object detection is one of the fundamental tasks in computer vision, requiring the model to return the category and location of each object in an image.
Generic object detection focuses on category-agnostic detection approaches rather than specific class-aware methods (e.g. face detection methods).
Compared to image classification, object detection not only needs to recognize the object but also demands localization.
Typically, the localization task is to regress the coordinates of the bounding box, which tightly bounds the target object.

\begin{figure}[!t]
	\centering
	\includegraphics[width=0.9\linewidth]{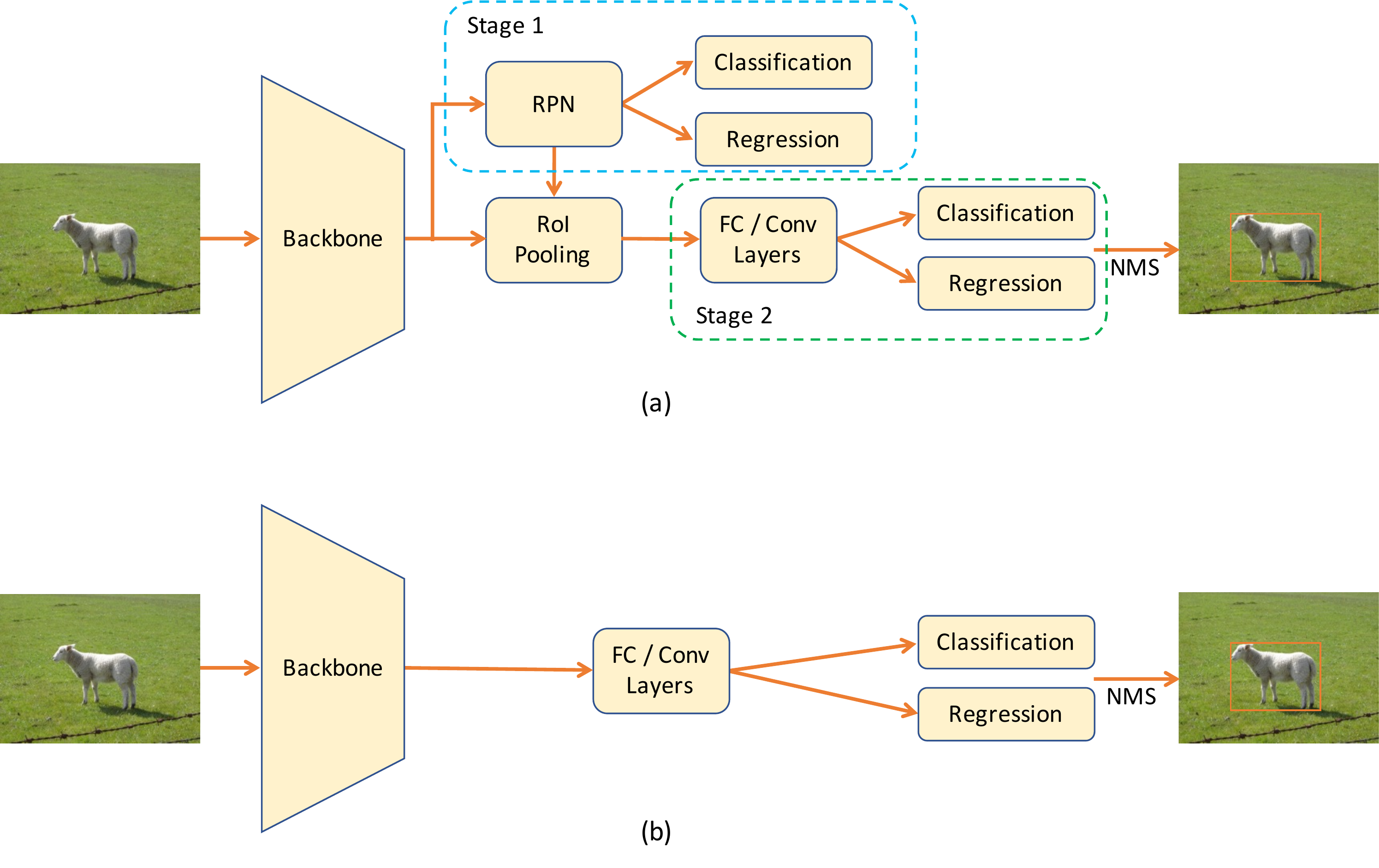}
	\caption{Standard frameworks of (a) two-stage object detection, and (b) one-stage object detection.}
	\label{fig:frcn}
\end{figure}

In the deep learning era, there are two mainstream object detection frameworks, i.e., two-stage detectors and one-stage detectors. Fig.~\ref{fig:frcn} shows the standard frameworks of two-stage detectors and one-stage detectors.
Two-stage detectors, such as R-CNN series~\cite{girshick2014rich, girshick2015fast, ren2015faster, he2017mask}, firstly generate category-agnostic region proposals by selective search~\cite{uijlings2013selective} or Region Proposal Network (RPN)~\cite{ren2015faster},
and then the proposals are projected onto the feature map through RoI (region of interest) Pooling/Align.
Finally, the proposal features are fed into two parallel branches to determine the class label and fine-tune the bounding box.
Comparatively, one-stage detectors like the YOLO series~\cite{redmon2016you, redmon2017yolo9000, redmon2018yolov3, bochkovskiy2020yolov4} and the SSD series~\cite{liu2016ssd, fu2017dssd, zhang2018single} directly predict the class possibilities and bounding box offsets for each spatial location of the full image without region proposals.
To remove the redundant detection results, a post-processing algorithm is applied in both frameworks, such as NMS~\cite{neubeck2006efficient} and Soft NMS~\cite{bodla2017soft}.
Generally, the advantage of two-stage detectors lies in high detection accuracy, while the advantage of one-stage detectors is fast speed.

On the other hand, the existing object detection methods can be grouped into two types: anchor-based and anchor-free.
The concept of anchor points was first proposed in Faster R-CNN~\cite{ren2015faster}, each of which indicates the predefined box with a specific aspect ratio and size,
providing RPN a good initialization to avoid too large search space and generating better region proposals.
In addition to two-stage detectors after Faster R-CNN, many one-stage detectors also utilize anchor points to improve the quality of proposals.
However, anchors bring many hyper-parameters and design choices via ad-hoc heuristics, and exacerbate the imbalance between positive and negative proposals as most anchors contain no objects.
CornerNet~\cite{law2018cornernet} proposes to detect objects as a pair of key points, i.e., the top-left corner and the bottom-right corner, which shows that detectors without anchors can obtain good results.
Subsequently, many anchor-free methods have been proposed and can be categorized into two types: keypoint-based and center-based ones.
Keypoint-based methods~\cite{law2018cornernet,duan2019centernet,zhou2019bottom, dong2020centripetalnet} detect a set of keypoints and match them to encode a bounding box.
Center-based methods~\cite{tian2019fcos, zhou2019objects, zhu2019feature, kong2020foveabox, zhang2020bridging} directly predict the center of an object and exploit different algorithms to define positive and negative samples.

Moreover, there are some approaches aiming to solve special problems in object detection.
FPN~\cite{lin2017feature} and its variants~\cite{tan2020efficientdet, liu2018path, ghiasi2019fpn} design different feature networks to strengthen the ability of detecting objects at various scales.
RetinaNet~\cite{lin2017focal} proposes the focal loss, which is now widely adopted to alleviate the foreground-background class imbalance issue.
Some methods such as FreeAnchor~\cite{zhang2019freeanchor}, Noisy Anchors~\cite{li2020learning}, PISA~\cite{cao2020prime}, and Guided Anchoring~\cite{wang2019region} focus on the label assignment problem, labeling positive anchors in a dynamic way rather than through a fixed IoU threshold.
DETR~\cite{carion2020end} presents a new method that treats object detection as a direct set prediction problem using Transformer~\cite{vaswani2017attention}, which creates a new paradigm for computer vision.
Subsequently, many Transformer based methods have been proposed to solve various computer vision tasks, such as ViT~\cite{dosovitskiy2020image}, Deformable DETR~\cite{zhu2020deformable}, and Swin Transformer~\cite{liu2021swin}.

\section{Overview of Few-Shot Object Detection}\label{sec:overview}
In this section, we first introduce the problem definition of FSOD and then introduce the common datasets and evaluation metrics, followed by a proposed taxonomy of FSOD works.

\subsection{Problem Definition}
FSOD aims at tackling the problem that there are only a few available samples for the classes to be detected.
Generally, there is a set of \textit{base classes} or \emph{seen classes}~$C_{base}$ with abundant labeled data for each class, and a set of \textit{novel classes} or \emph{unseen classes}~$C_{novel}$, each of which has only a few annotated samples.
It is worth noting that~$C_{base}\cap C_{novel}=\emptyset$.
Similar to few-shot image classification (FSC), we consider an \textit{N}-way \textit{K}-shot problem (\textit{K} is usually no more than 30), which means $\vert C_{novel}\vert =$\textit{N} and each class in $C_{novel}$ contains \textit{K} instances.
The \textit{N}$\times$\textit{K} images constitute the \textit{support set}, and all the query images form the \textit{query set Q}.
Samples of the \textit{base classes} make up the \textit{base set}.

Given the \textit{base set} $B$ and the \textit{support set} $S$, and assume that the oracle dataset of the \textit{novel classes} is $D$, FSOD aims to optimize the model parameters $\theta$ learned from $B$ and $S$:
\begin{equation}
\label{eq_def}
\begin{aligned}
	\theta = \underset{\theta}{\text{arg min}} \sum_{(x_i,c_i,t_i)\in D} & (\mathcal{L}^{cls}(\mathrm{CLS}(x_i),c_i;\theta^{B,S})\\
	&+\mathcal{L}^{loc}(\mathrm{LOC}(x_i),l_i;\theta^{B,S}))\\
	s.t. \quad c_i\in C_{novel}
\end{aligned}
\end{equation}
where $\mathcal{L}^{cls}$ and $\mathcal{L}^{loc}$ are the classification and regression loss function, respectively. $x_i$ indicates an object of a certain \textit{novel class}, and $c_i$ and $l_i$ are its class label and location.

\begin{table}
  \begin{center}
  \caption{Statistic comparison among common few-shot object detection (FSOD) datasets.}
  \label{tab:datasets}
  \resizebox{\linewidth}{!}{
  \begin{tabular}{ c  c  c  c  c }
  \toprule
  Dataset & \#Base Classes & \#Novel Classes & \textit{K} values & Remark\\
  \midrule
  PASCAL VOC~\cite{everingham2010pascal, everingham2015pascal, kangFewShot2019} & 15 & 5 & 1,2,3,5,10 & 3 different base/novel splits\\
  MS-COCO~\cite{lin2014microsoft, kangFewShot2019} & 60 & 20 & 10,30 & novel classes overlap with VOC classes\\
  \bottomrule
  \end{tabular}}
  \end{center}
\end{table}

\subsection{Datasets}
In the beginning, each FSOD method has its own data setting and dataset.
Kang et al.~\cite{kangFewShot2019} later proposed standard splits on generic object detection datasets (Kang's split), i.e., PASCAL VOC~\cite{everingham2010pascal, everingham2015pascal} and MS-COCO~\cite{lin2014microsoft}, which have been accepted as the most common benchmarks for FSOD.

\begin{itemize}
\item \textbf{PASCAL VOC (Kang's split)}
There are two versions of PASCAL VOC, i.e., VOC07~\cite{everingham2010pascal} and VOC12~\cite{everingham2015pascal}. Kang et al.~\cite{kangFewShot2019} used VOC07 and VOC12 train/val sets for training and VOC07 test set for testing.
Out of the 20 categories in total, 5 random classes are chosen as novel classes, while the remaining 15 classes are kept as base classes.
To make the evaluation more convincing, Kang et al. generated 3 different base/novel splits.
To construct the \textit{N}-way \textit{K}-shot problem, Kang et al.~\cite{kangFewShot2019} suggested to set \textit{K} $\in\{1, 2, 3, 5, 10\}$.

\item \textbf{MS-COCO (Kang's split)}
Similar to PASCAL VOC dataset, 5000 images from the MS-COCO~\cite{lin2014microsoft} validation set (called \textit{minival}) are used for testing, and the rest of the images in the train/val sets are used for training.
It is worth noting that in MS-COCO2017, the training set equals to the combination of the training and validation set without \textit{minival} of MS-COCO2014, and the validation set equals to the \textit{minival} set.
There are 80 categories in the MS-COCO dataset, in which 20 categories overlapping with the PASCAL VOC dataset are chosen as novel classes, and the remaining 60 categories are used as base classes.
Each novel class has \textit{K} labeled image, where \textit{K} $\in\{10, 30\}$.
\end{itemize}
%

\begin{figure}[!t]
	\centering
	\includegraphics[width=0.95\linewidth]{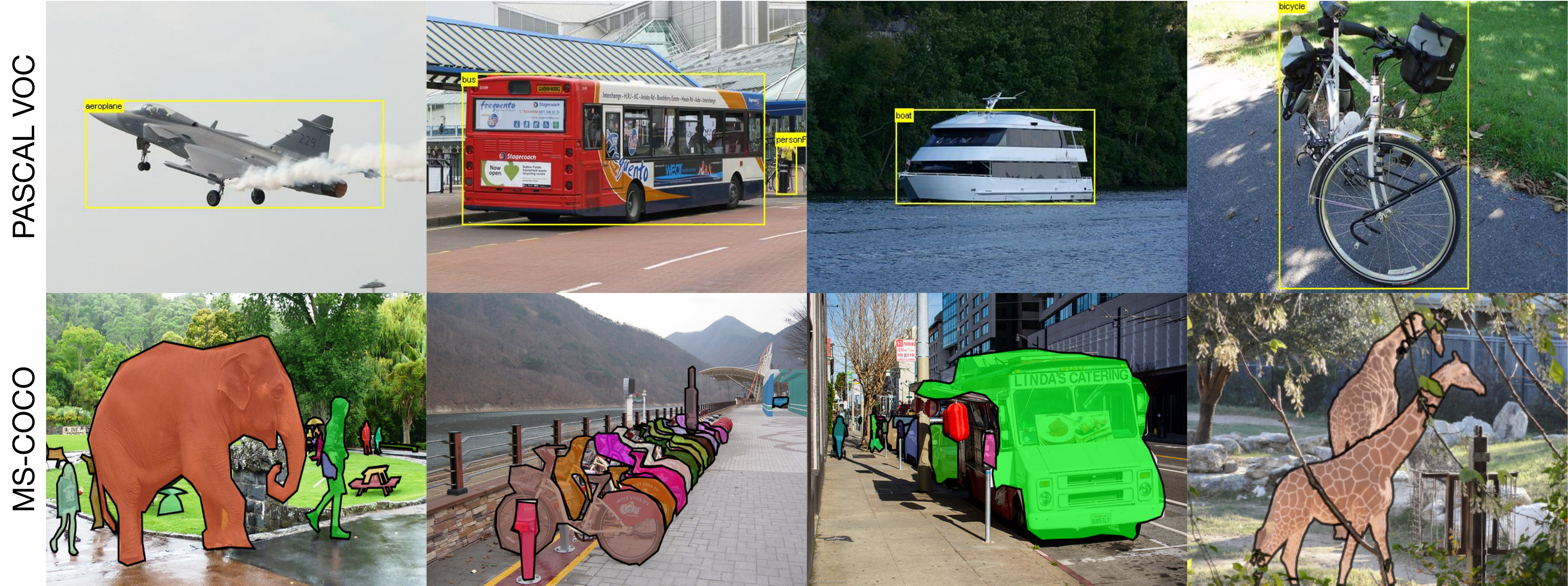}
	\caption{Examples of annotated images in two popular few-shot object detection datasets: PASCAL VOC and MS-COCO. We present four samples of each dataset.
	}
	\label{fig:dataset_example}
\end{figure}

The comparison among the FSOD datasets above is summarized in Table~\ref{tab:datasets}.
Fig.~\ref{fig:dataset_example} illustrates some examples in the datasets.

\subsection{Evaluation Metrics}
In object detection, mean Average Precision (mAP) is a standard evaluation metric introduced by VOC07~\cite{everingham2010pascal}.
AP is defined as the average precision under different recalls for each category, and mAP is the mean of AP over all categories.
To measure the localization accuracy of each prediction box, the Intersection over Union (IoU) is used, which represents the overlap ratio between two boxes.
Concretely, if the IoU between the ground-truth box and the prediction box is bigger than a predefined threshold, the prediction box is treated as a true positive, otherwise, it is treated as a false positive.
In PASCAL VOC, a fixed IoU threshold of 0.5 is used.
Differently, MS-COCO~\cite{lin2014microsoft} uses multiple IoU thresholds between 0.5 and 0.95 to encourage more accurate object detection methods.
In addition, MS-COCO reports the mAP on small, medium, and large objects, respectively.

Similar to FSC, early FSOD methods focus only on the performance on novel classes.
Kang et al.~\cite{kangFewShot2019} pointed out that we should not only report the performance on novel classes, but also pay attention to the performance on base classes to provide a more comprehensive evaluation.
Later works report the performance on both base classes and novel classes, which are denoted as bAP and nAP, respectively.

\begin{figure}[!t]
	\centering
	\includegraphics[width=1\linewidth]{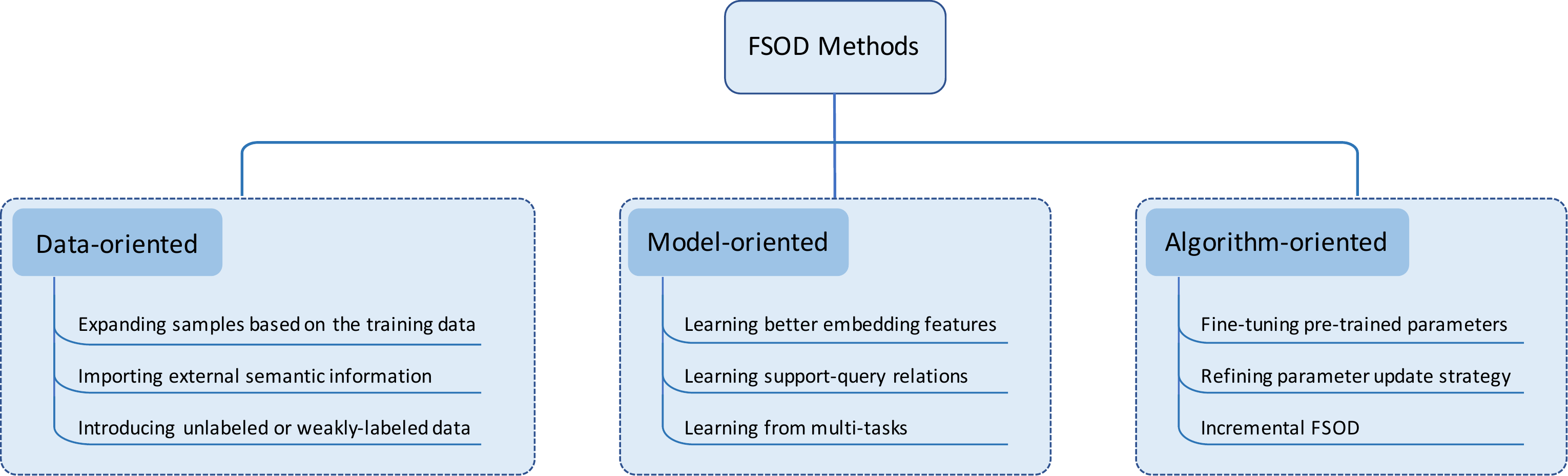}
	\caption{Our proposed taxonomy of existing FSOD methods.
      We categorize the existing FSOD methods into three types: data-oriented, model-oriented, and algorithm-oriented, and each type is further classified into several subgroups.}
	\label{fig:taxonomy}
\end{figure}

\subsection{Our Taxonomy of FSOD}
In this paper, we categorize existing FSOD methods according to their major contribution from three perspectives: data, model, and algorithm.
In our point of view, each work contributes to the area in a major aspect: data, model, or algorithm.
Therefore, different from simply classifying the existing FSOD methods into fine-tuning-based ones and meta-learning-based ones, here we propose a rational taxonomy of the existing FSOD methods that fall into three types: data-oriented, model-oriented, and algorithm-oriented as shown in Fig.~\ref{fig:taxonomy}:
\begin{itemize}
  \item \textbf{Data-oriented}. Challenges of FSOD are mainly caused by the limited data, thus it is an intuitive way to expand the training samples. Data-oriented methods try to overcome the limitation of training sample shortage of novel classes by generating more samples or providing additional information for model training.
  Besides conventional data augmentation strategies, some methods develop specific advanced data generation modules or import external semantic information from natural language.
  \item \textbf{Model-oriented}. Methods in this type mainly modify the existing generic object detection frameworks to adapt to the few-shot scenario.
  The main idea is to effectively separate different classes in the feature space with the help of only a few given samples.
  Some dedicate to designing better modules to generate better embedding features, while some others choose to mine the relationship between the support and query images.
  And there are also some methods combining other related tasks with FSOD, which is able to simultaneously improve the performance of all tasks.
  \item \textbf{Algorithm-oriented}.
  Instead of carefully designing the detection framework, methods in this type concentrate mainly on parameter initialization or update strategy with the help of pre-trained parameters.
  Moreover, some specific approaches propose special algorithms to solve the incremental few-shot object detection problem that demands detecting unseen classes without re-training and remembering seen classes at the same time.
\end{itemize}

Note that Wang et al.~\cite{wang2020generalizing} categorized FSL from three perspectives: data, model, and algorithm. The main differences between their survey and ours lie in two aspects:  on the one hand, they tried to differ the methods from how to search for the best hypothesis in the hypothesis space with the help of prior knowledge, while we aim to subsume the methods in terms of their main contributions. On the other hand, they focused on the object classification task, while we concentrate on the object detection problem.

In the following section, we will elaborate on each type of FSOD method in detail.

\section{Recent Advances in Few-Shot Object Detection}\label{sec:fsod}
In this section, we present a detailed review on the achievements of existing FSOD works, which are grouped into three types: data-oriented, model-oriented, and algorithm-oriented. And each type of them is further divided into fine-grained subgroups.

\subsection{Data-oriented Methods}
\begin{figure}[!t]
	\centering
	\includegraphics[width=0.9\linewidth]{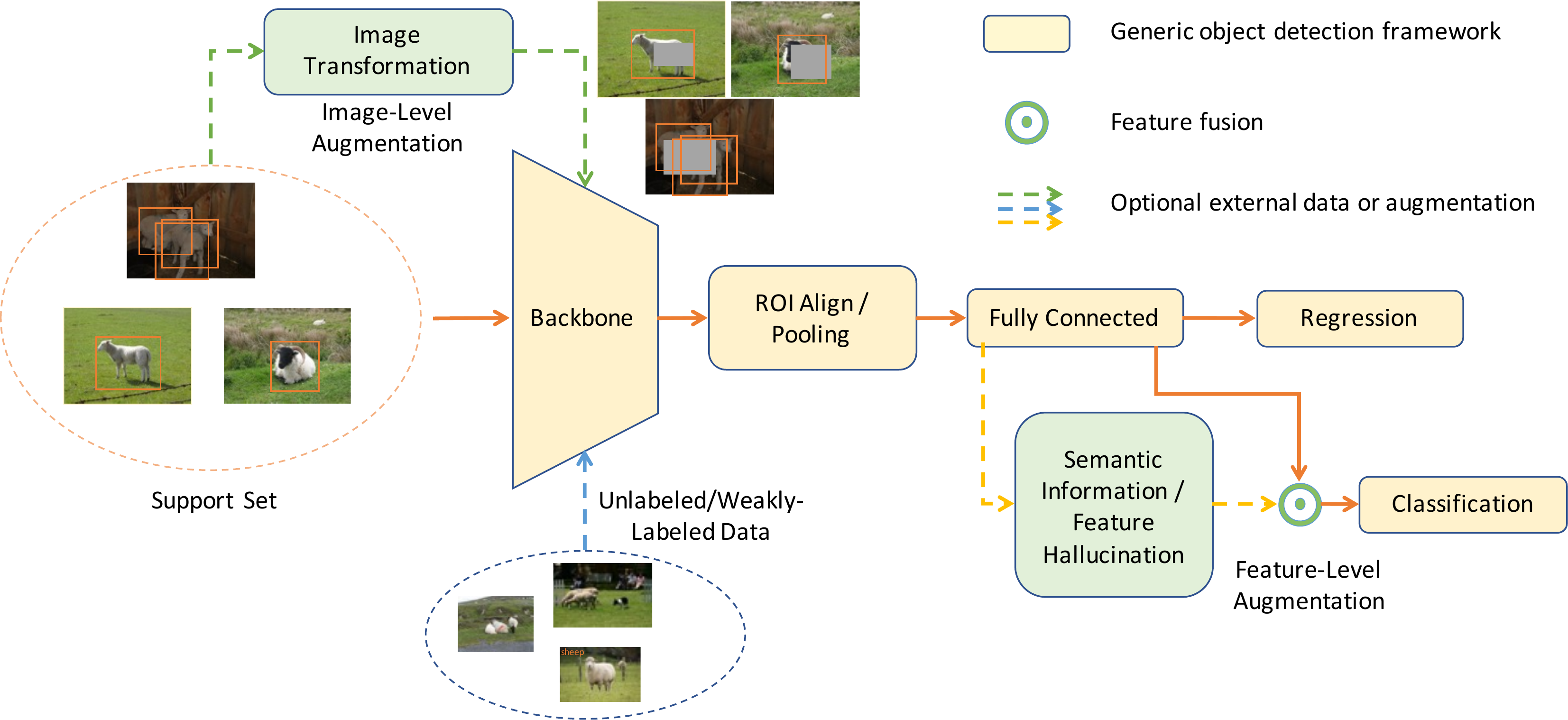}
	\caption{The major schemes of data-oriented few-shot object detection methods.
      Some methods learn to generate novel image-level (image transformation) or feature-level (feature hallucination) samples based on the \textit{support set},
      while others obtain extra prior knowledge from unlabeled/weakly-labeled images or semantic information that is learned from a large corpus.}
	\label{fig:data-oriented}
\end{figure}

Methods in this type aim at augmenting the training samples or information of novel classes.
Some methods focus on extending the \textit{support set} through specific data augmentation approaches, while some import external semantic information to assist the classification task in FSOD.
To further mitigate the labeling effort, some other models utilize unlabeled or weakly-labeled data.
Fig.~\ref{fig:data-oriented} illustrates the major schemes of existing data-oriented FSOD methods.
Assume that $E$ indicates the extra data or augmented support data, Equ.~(\ref{eq_def}) can be modified as follows:
\begin{equation}
\label{eq_data}
\begin{aligned}
	\theta = \underset{\theta}{\text{arg min}} \sum_{(x_i,c_i,t_i)\in D} & (\mathcal{L}^{cls}(\mathrm{CLS}(x_i),c_i;\theta^{B,S,E})\\
	&+\mathcal{L}^{loc}(\mathrm{LOC}(x_i),l_i;\theta^{B,S,E}))\\
	s.t. \quad c_i\in C_{novel}
\end{aligned}
\end{equation}

\subsubsection{\textbf{Expanding samples based on the training data}}
One intuitive way to FSOD is to expand the training samples by transforming the support set.
Although traditional data augmentation methods such as flipping and cropping are widely used in object detection, they could only provide limited data variation.
Li et al.~\cite{liTransformation2021} showed that simply adding naive data augmentation techniques is not helpful for FSOD, and
they pointed out that the reason lies in the guidance inconsistency between different transformed images.
To tackle this issue, they presented a new consistency regularization method to constrain the features of various transformed images, which was implemented by a new loss function.
Moreover, a proposal detection network was designed to generate the transformation invariant features.

In addition to image-level augmentation, feature-level expansion is also considered an effective way.
Zhang et al.~\cite{zhangHallucination2021} proposed a hallucinator module in the RoI feature space.
As many modes of visual changes such as illumination and camera pose are category-agnostic, the hallucinator trained on base classes can be generalized onto novel classes to generate additional novel class examples.
In order to mitigate the difficulty of applying the hallucinator to existing FSOD methods, they adopted an \textit{EM-like} training strategy.

\subsubsection{\textbf{Importing external semantic information}}

Intra-class visual information is very variant due to different textures, shapes, sizes, etc.,
thus in the few-shot scenario, the lack of samples hurts the detection performance.
Oppositely, semantic information is usually invariant for each category.
Besides, it is easier to construct explicit inter-class relationships.

SRR-FSD~\cite{zhuSemantic2021} utilizes a word embedding network~\cite{mikolov2013distributed} along with a relation reasoning module to expand the prior knowledge.
The word embedding network is pre-trained on a large corpus of text.
The visual features and corresponding class semantic embeddings are aligned to obtain enhanced representations, leading to better classification results.
FADI~\cite{cao2021few} presents a new idea that firstly associates each novel class with a base class, and then discriminates them.
The association step adopts WordNet~\cite{miller1995wordnet}, an English vocabulary graph that incorporates rich lexical knowledge, to describe class similarities.
The class-to-class similarity is further measured by Lin Similarity~\cite{lin1998information}.
And each novel class is given a pseudo base class label during training.
In the discrimination step, the classification branches for base and novel classes are disentangled.
In such a way, the distance within each novel class is shrunk while the distance between novel classes is enlarged, leading to better detection performance.

\subsubsection{\textbf{Introducing unlabeled or weakly-labeled data}}

Although instance-level annotation is labor-intensive and time-consuming, unlabeled and image-level labeled data are much easier to obtain.
Semi-supervised object detection~\cite{cheng2014semi, tang2016large, tang2017visual, gao2019note, jeong2019consistency, wang2021data} and weakly-supervised object detection~\cite{bilen2014weakly, bilen2015weakly, zhu2017soft, tang2018pcl, wan2018min, ren2020instance, dong2021boosting} have been developed for a long time, but most of these methods still require much more data than FSOD methods.
Therefore, some specific semi-supervised or weakly-supervised FSOD methods were proposed for FSOD.

Dong et al.~\cite{dongFewExample2019} proposed a semi-supervised object detection framework that needs only 2-4 annotated objects for each class.
The proposed model MSPLD iterates between training and high-confidence sample generation, during which challenging but reliable samples are generated and the detector becomes more robust.
In addition, multiple detectors are integrated to improve both precision and recall.
Hu et al.~\cite{huSILCO2019} presented a weakly-supervised FSOD framework named SILCO, which aims to detect objects in the query image with a few weakly-supervised labeled support images.
A spatial similarity module is designed to extract the spatial commonality among the query and support images, and a feature re-weighting module is proposed to measure the importance of different support images via graph convolutional networks.
Another weakly-supervised object detection method StarNet~\cite{karlinskyStarNet2020} utilizes a non-parametric star-model to achieve geometric matching for each query-support image pair.
It adopts heatmaps to indicate where the query and support image matches. During training, the matched regions are gradually highlighted, thus objects in the query image are detected.
As a weakly-supervised object detection framework, StarNet is easy to be applied to few-shot classification.
MINI~\cite{cao2022mini} aims to mine implicit novel objects in base images due to possible co-occurrence of base and novel objects.
It utilizes semi-supervised learning to annotate implicit novel instances as pseudo labels via adaptive mingling of both offline and online mining mechanisms.
Kaul et al.~\cite{kaul2022label} designed a simple training strategy to reduce missing detection.
Specifically, they imported a kNN classifier using the output class token of a ViT~\cite{dosovitskiy2020image} model pre-trained on a large number of unlabeled images with a self-supervised method DINO~\cite{caron2021emerging}.
The model trained on the novel set predicts novel objects on the training set, and the kNN classifier was used to correct the results.
Finally, three class-agnostic separated regressors were utilized to further refine the boxes.

\begin{table}[]
\caption{A qualitative comparison among data-oriented FSOD methods. Note that SILCO~\cite{huSILCO2019} and StarNet~\cite{karlinskyStarNet2020} only utilize weakly-labeled images for training.}
\label{tab:data-oriented}
\resizebox{\linewidth}{!}{
\begin{tabular}{cccccc}
\toprule
  Category & Model  & Year & Extra Visual Data & Extra Semantic Data                                \\
\bottomrule
\multirow{2}{*}{Expanding samples based on the training data} & TIP~\cite{liTransformation2021}     & 2021 & image-level augmentation & -  \\
    & Halluc~\cite{zhangHallucination2021}  & 2021 & feature-level augmentation & -  \\
\midrule
\multirow{2}{*}{Importing external semantic information}      & SRR-FSD~\cite{zhuSemantic2021} & 2021 & - & word embedding vector of each class &                      \\
    & FADI~\cite{cao2021few}    & 2021 & - & class-to-class similarity via WordNet~\cite{miller1995wordnet} \\
\midrule
\multirow{5}{*}{Introducing unlabeled or weakly-labeled data} & MSPLD~\cite{dongFewExample2019}   & 2019 & unlabeled images & - \\
    & SILCO~\cite{huSILCO2019}   & 2019 & weakly-labeled images & -  \\
    & StarNet~\cite{karlinskyStarNet2020} & 2020 & weakly-labeled images & -   \\
    & MINI~\cite{cao2022mini} & 2022 & implicit novel objects in base images & -   \\
    & Kaul et al.~\cite{kaul2022label} & 2022 & output class token of ViT~\cite{dosovitskiy2020image} & -   \\
\bottomrule
\end{tabular}}
\end{table}

\subsubsection{\textbf{Discussion and Conclusion}}

Data augmentation is a simple way to improve the detection performance.
However, expanding training samples based on the \textit{support set} mainly enhances the model's ability of anti-interference, but cannot expand the data distribution.
Importing other data such as semantic information or extra unlabeled data is helpful to expand the diversity of data, but it is unfair compared to the standard FSOD methods.
There is still a need for methods that can generate more widely distributed data.
Differences between data-oriented FSOD methods are shown in Table~\ref{tab:data-oriented}.

\subsection{Model-oriented Methods}
FSOD methods in this type focus on modifying the generic detection framework to effectively distinguish different classes based on a few given samples.
Some methods dedicate to generating discriminative features for each class, while some other methods learn the support-query relations to decrease the demand for training samples.
Moreover, multitask learning is also an effective way to improve FSOD performance.

\subsubsection{\textbf{Learning better embedding features}}
\begin{figure}[!t]
  \includegraphics[width=0.9\linewidth]{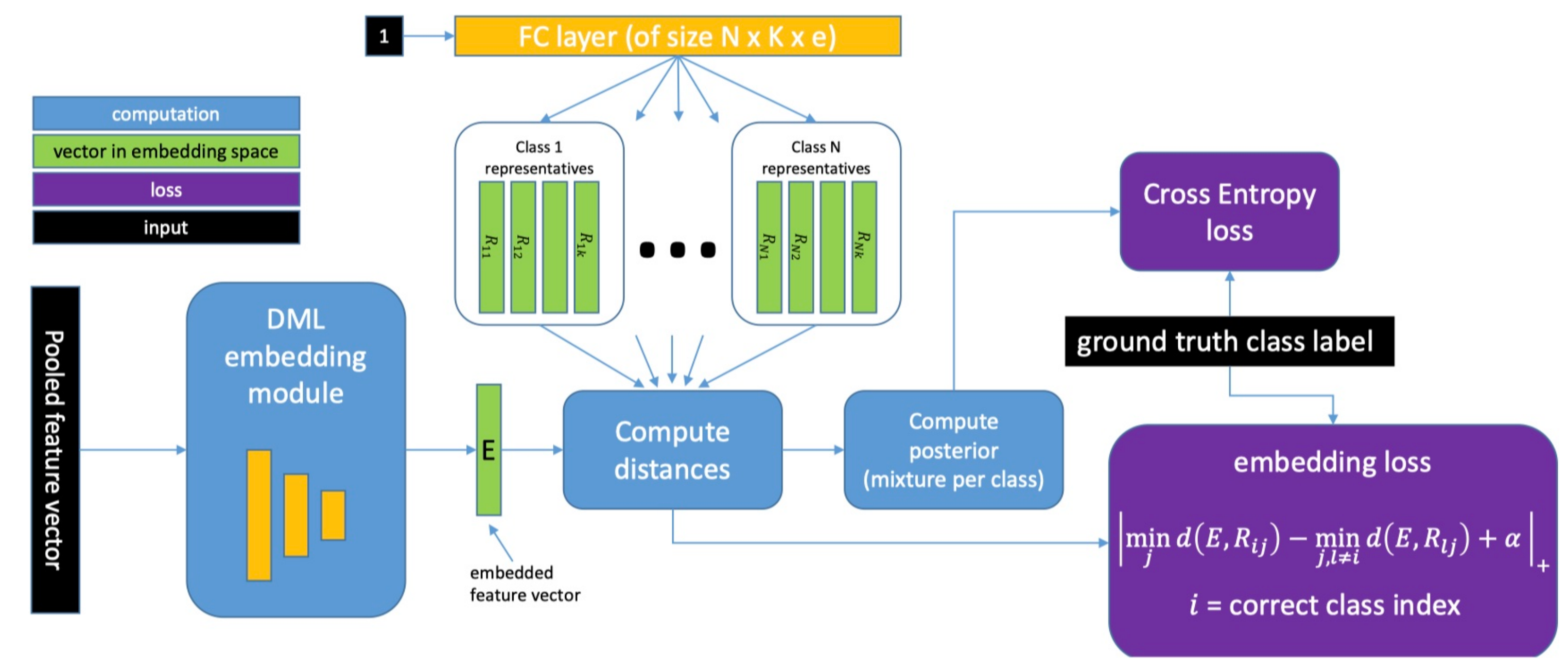}
  \caption{RepMet~\cite{karlinskyRepMet2019} architecture. Each class is represented by $K$ vectors obtained via a fully connected (FC) layer. The output of RoI Pooling is fed into an embedding module to obtain the embedded feature vector, which is used to compute distances between the class representatives, thus the query proposal is classified as the closest class. The embedding module is supervised by the embedding loss. Image courtesy of RepMet~\cite{karlinskyRepMet2019}.}
  \label{fig:repmet}
\end{figure}

For methods of this type, a widely adopted strategy is to learn an adequately distinctive embedding feature in a lower-dimensional space for each class.
In such a way, the intra-class distance is shrunk while the inter-class distance is enlarged.
Therefore, the classes are separated better with a few samples and the detection performance is improved.
The classification branch $\mathrm{CLS}(x_i)$ in Equ.~(\ref{eq_def}) can be extended as follows:
\begin{equation}
\label{eq_embed}
\begin{aligned}
	\mathrm{CLS}(x_i;S)&=\underset{c_i}{\text{arg min}}~\mathcal{D}(\mathcal{F}(x_i), \mathcal{G}(s_i))\\
	s.t. \quad &(s_i,c_i)\in S
\end{aligned}
\end{equation}
where $\mathcal{D}$ indicates the distance function that measures the similarity between the query proposal features and the support class embedding features or prototype features.

Some methods learn the features under the guidance of a specially designed metric between two samples, so we call them \emph{metric-based methods}.
A representative is RepMet~\cite{karlinskyRepMet2019}, which proposes a novel distance metric learning framework for both FSC and FSOD, shown in Fig~\ref{fig:repmet}.
Each class is represented by a set of vectors, and the vectors contain \textit{K} modes, as RepMet assumes there are \textit{K} peaks in the distribution of each class.
An embedding loss is designed to guarantee that the distance between the embedding feature and the closest correct class representative is at least $\alpha$ shorter than the distance between the embedding feature and the closest wrong class representative, where $\alpha$ is a hyperparameter.
NP-RepMet~\cite{yangRestoring2020} improves RepMet~\cite{karlinskyRepMet2019} via restoring the negative information as it is important as well to embedding space learning.
The class representative module and the embedding vector module are all divided into two branches for positive and negative proposals, respectively.
Given a positive embedding vector, two distances are computed separately, i.e., the closest distance from it to the positive class representatives and the closest distance from it to the negative class representatives.
A modified triplet loss~\cite{schroff2015facenet} is defined to optimize the embedding space.
DMNet~\cite{lu2022decoupled} solves the problem that RepMet~\cite{karlinskyRepMet2019} is not capable of handling the classification of multiple objects through designing an image-level parallel distance metric learning method. Moreover, to further improve the single-stage detector performance, DMNet decouples classification and localization into two branches and proposes a new module called decoupled representation transformation, which predicts objectness and anchor shapes to alleviate the adverse effect of handcrafted anchors.

Several methods adopt the idea of Prototypes Networks~\cite{snell2017prototypical} to learn the prior knowledge of each class from the support set. We call them \emph{prototype-based methods}.
Meta-RCNN~\cite{wuMetaRCNN2020} learns the region features via Prototypical Networks~\cite{snell2017prototypical} to construct the class prototypes,
and based on which a class-aware attention map is used for enhancing the target class features.
The learned prototypes are finally compared with each query RoI feature to filter the positive detection results.
FSOD$^{up}$~\cite{wuUniversalPrototype2021} claims that it is better to learn universal prototypes across different categories rather than learn class-specific prototypes.
The prototypes are learned in two steps:
1) the universal prototypes are learned to represent the image-level information;
2) After RPN, the universal prototypes are transformed into conditional prototypes that mainly include object-level information.
The object features are finally enhanced with a soft-attention of the learned prototypes.
Different from the aforementioned methods, Lee et al.~\cite{leeFewShot2021} devised to treat each support image as an individual prototype.
To refine the support prototypes, an attention mechanism was proposed to extract the shared information among them.
Then, another attention module was introduced that leverage each support prototype as a class code by comparing them with the query features.
Han et al.~\cite{han2021meta} proposed a Meta-Classifier to measure the similarity between the noisy proposals and prototypes of few-shot novel classes.
Concretely, soft correspondences between each spatial position in the proposal features and class prototypes are first calculated, and the spatial alignment is then performed.
The aligned features are finally fed into a prototype matching network to obtain the similarities between them and novel classes prototypes.

\begin{figure}[!t]
	\centering
	\includegraphics[width=0.8\linewidth]{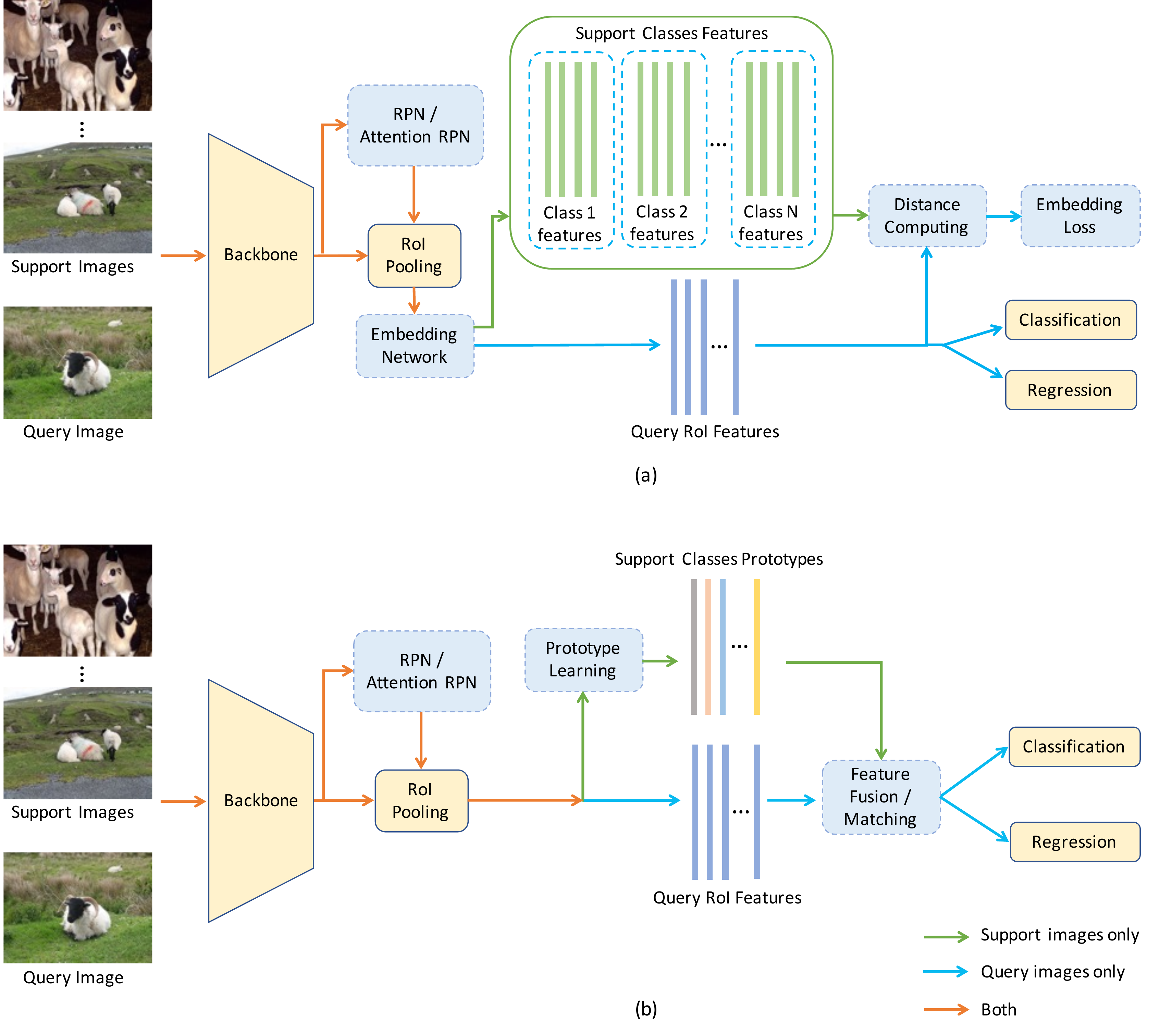}
	\caption{Comparison between metric-based and prototype-based FSOD methods. (a) Metric-based methods learn the features via a specially designed distance computing module; (b) Prototype-based methods try to learn class prototypes, which are used to enhance the query features or compare with the query features.}
	\label{fig:embedding}
\end{figure}

To better illustrate the difference between metric-based and prototype-based FSOD methods, in Fig.~\ref{fig:embedding} we show the architectures of metric-based and prototype-based FSOD methods.

Some other methods focus on generating more discriminative class-level features.
CME~\cite{liMaxMargin2021} intends to tackle the contradiction between novel class classification and representation through a class margin equilibrium method.
Concretely, during base training, a class margin loss is introduced to maximize the distances between novel classes.
Then during fine-tuning, a feature disturbance strategy that truncates gradient maps is proposed.
In this way, an adversarial learning procedure for min-max margin is defined to achieve class margin equilibrium.
Wu et al.~\cite{wu2021generalized} utilized singular value decomposition (SVD) to enhance both the generalization and discrimination abilities.
Specifically, localization is performed in the generalization space, which is built by eigenvectors with larger singular values,
while classification is performed in the discrimination space, which is built through eigenvectors with smaller singular values.
To further improve the classification performance, dictionary learning~\cite{zhang2017deep} is adopted, which is capable of capturing high-level discriminative information.
Li et al.~\cite{liFewShot2021a} proposed a novel refinement mechanism to boost classification performance by plugging an additional discriminability enhancement branch to the original Faster R-CNN~\cite{ren2015faster}.
During training, the extra branch utilizes the misclassified false positives of the base detector to improve the ability of reducing category confusion.
Wu et al.~\cite{wu2022learning} solved the FSOD problem from the perspective of suppressing imbalanced data distribution between base and novel classes. 
Specifically, the distribution calibration module was devised to balance the distributions and decision areas of base and novel classes, and the class discrimination regularization module was proposed to enlarge the distance between different classes.
HiCLPL~\cite{zhang2022hierarchical} proposes a new problem called hierarchical few-shot object detection, which aims to detect objects with hierarchical categories under the few-shot setting.
A hierarchical contrastive learning approach is developed to constrain the class feature spaces to ensure consistency with the hierarchical taxonomy.
Moreover, a probabilistic loss is designed to make child nodes able to correct the classification errors of their parent nodes in the taxonomy.

Some methods concentrate on producing more differential instance-level features.
Context-Transformer~\cite{yangContexttransformer2020} leverages the context of each prior box to reduce the object confusion.
The detector is trained on the source domain and then fine-tuned on the target domain, aiming to generate reliable prior boxes.
The surrounding information of each prior box is treated as the contextual field,
and an affinity discovery module is proposed to construct relations between the prior boxes and the contextual fields.
Then, a context aggregation module integrates context into the representation of each prior box.
FSCE~\cite{sunFSCE2021} also observes that the detection error is mainly from misclassifying novel instances into other confusable classes.
It treats region proposals of different IoUs for an object as intra-image
augmentation croppings~\cite{khosla2020supervised}.
Therefore, to learn a well-separated decision boundary, the supervised contrastive approach is extended to ease the misclassification problem.
The original RoI head in Faster R-CNN~\cite{ren2015faster} is augmented with a new branch that measures the similarity between RoI features,
and a contrastive proposal encoding loss is defined to increase the similarity between proposal features with the same class label
while enlarging the distance between proposal features with different labels in the embedding space.
Kim et al.~\cite{kim2022few} first analyzed and addressed the fundamental weaknesses of the classical fine-tuning-based pipeline, where optimizing the detector to the novel RoIs obeys the severely imbalanced IoU distribution.
Then, they proposed a novel approach that utilizes three detection heads with different parameters to balance out the disproportionate IoU distribution and enrich the RoI samples to boost the performance.

MPSR~\cite{wuMultiscale2020} tackles the FSOD problem from the perspective of dealing with scale variations.
Due to the sparse scale distribution in the few-shot setting, it is more challenging to detect objects of different scales.
MPSR achieves multi-scale detection by generating positive samples of different scales as object pyramids.
To avoid introducing improper negatives, MPSR adaptively selects specific features from the feature maps in the FPN~\cite{lin2017feature} stage as positives for each object.

\subsubsection{\textbf{Learning support-query relations}}
Along this line, FSOD methods are dedicated to fully exploiting support features through mining the relationship between the support and query images and then guiding the representations of the query features.
The mined relationship is helpful for better classification.
Here, the classification branch $\mathrm{CLS}(x_i)$ can be written as:
\begin{equation}
\label{eq_relation}
\begin{aligned}
	\mathrm{CLS}(x_i;S)&=\mathrm{CLS}(\mathrm{Relation}(\mathcal{F}(x_i),\mathcal{F}(s_i))) \\
	s.t. \quad &s_i\in S
\end{aligned}
\end{equation}

\begin{figure}[!t]
	\centering
	\includegraphics[width=0.9\linewidth]{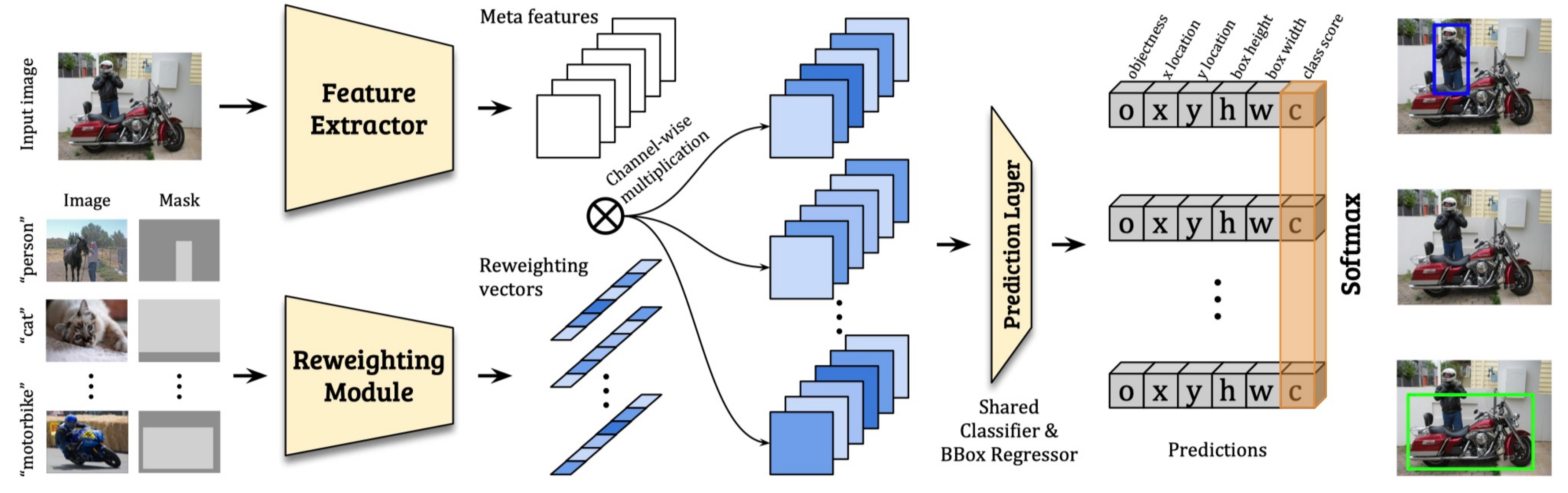}
	\caption{Model architecture of FSRW~\cite{kangFewShot2019}. It follows the one-stage detector framework, where a meta-feature extractor directly regresses and classifies objects in the query image, obtaining the objectness score ($o$), bounding box location ($x,y,w,h$), and class score (c). The re-weighting module is proposed to extract global features of each support image belonging to $N$ classes and map them into $N$ re-weighting vectors, which are used to modulate the meta-features for better prediction. Image courtesy of FSRW~\cite{kangFewShot2019}.}
	\label{fig:fsrw}
\end{figure}

FSRW~\cite{kangFewShot2019} is a classical method that addresses the standard FSOD task and presents an efficient meta-learning-based FSOD framework.
The architecture is shown in Fig.~\ref{fig:fsrw}.
Taking YOLO-v2~\cite{redmon2017yolo9000} as a meta-feature extractor, a novel feature re-weighting module acquires the global feature of each support image and generates re-weighting vectors.
The vectors are utilized as coefficients to refine the query meta-features.
Therefore, support-class-specific information is integrated into the query features for better representations.

Most methods adopt the attention mechanism for support-query relation mining, which is simple yet efficient to highlight the most important parts of the input images.
Hsieh et al.~\cite{hsiehOneShot2019a} devised a novel co-attention and co-excitation framework to tackle the one-shot object detection problem.
The non-local operation~\cite{wang2018non} was used to extract the co-attention visual cues of both the support and query images.
Inspired by SENet~\cite{hu2018squeeze}, a novel squeeze-and-co-excitation technique was proposed to further explore the relationship between the support and query images.
Attention-RPN~\cite{fanFewShot2020} adopts three different relation networks to measure the similarity between two features,
where the relation networks represent the global, local, and patch relation respectively.
The original RPN in Faster R-CNN~\cite{ren2015faster} is replaced by the attention RPN, intending to integrate the support information.
Instead of extracting the global feature of each support image, DCNet~\cite{huDense2021} matches support and query features at a pixel-wise level.
Specifically, for each support image and each query image, a key map and a value map are generated to encode visual information.
The proposed dense relation distillation module densely matches the key maps between the query and support features, thus the query feature is enhanced with weighted support features.
The context-aware feature aggregation module is devised to capture multi-scale features with three different pooling resolutions.
DAnA~\cite{chenDualAwareness2021} focuses on the spatial misalignment and vagueness problem of feature representation.
The proposed method includes two attention modules.
One is the background attenuation block, which suppresses the background features.
The other is the cross-image spatial attention block, transforming the support features into query-position-aware features.
Zhang et al.~\cite{zhangAccurate2021} proposed a support-query mutual guidance mechanism, which not only considers enhancing the query features guided by the support set but also aggregates the support features guided by the query image.
Specifically, a support weighting module aggregates the support features according to the similarities between the support and query features.
Subsequently, the dynamic convolution~\cite{chen2020dynamic} that generates support-specific kernels was used to extract the query features.
KFSOD~\cite{zhang2022kernelized} leverages kernelized matrices to generate non-linear representation for each support region. The support representations are cross-correlated against the query image features to obtain attention weights, which are used to generate query proposals. A new module called multi-head relation net finally learns the support-query relations by combining all kernelized representations with the original feature maps.
To fully exploit the relations between support and query images, Park et al.~\cite{park2022hierarchical} designed the hierarchical attention module that extends the receptive field and computes the global attention of the query image. Moreover, meta-contrastive learning was utilized to increase the similarity between support and query objects belonging to the same class.

As a special type of attention-based framework, Transformer~\cite{vaswani2017attention} has shown its advantage in computer vision tasks~\cite{carion2020end, dosovitskiy2020image, zhu2020deformable, liu2021swin} recently, and some methods introduce Transformer to the FSOD problem.
Chen et al.~\cite{chenAdaptive2021} proposed the notion of language translation to one-shot object detection and implemented it based on Transformer~\cite{vaswani2017attention}.
First, they designed a multi-head co-attention module to correlate the support and query images.
Then, an adaptive image transformer module was employed to extract common semantic attributes for each support-query pair.
Finally, a selective channel attention was used to effectively fuse the information of all heads.
Another Transformer based FSOD method Meta-DETR~\cite{zhangMetaDETR2021} discards conventional Faster R-CNN framework and constructs an entire image-level detection model based on Deformable DETR~\cite{zhu2020deformable}.
The query features and multiple support classes are aggregated by a novel correlational aggregation module,
which is able to pay attention to all support classes simultaneously, instead of one-class-by-one-class detection with repeated runs on one query image as in most meta-learning methods.
Similar to Meta-DETR~\cite{zhangMetaDETR2021}, FS-DETR~\cite{bulat2022fs} builds a few-shot object detector based on Conditional DETR~\cite{meng2021conditional} and is capable of detecting multiple novel objects at once, supporting a variable number of samples per class.
To achieve testing novel classes without re-training, FS-DETR~\cite{bulat2022fs} treats support features as visual prompts, which are encoded as pseudo classes.
FCT~\cite{han2022fct} incorporates the cross-transformer into Faster R-CNN~\cite{ren2015faster} for not only the detection head but also the feature backbone to encourage multi-level interactions between the query and support, which helps the model improve the support-query relation mining.
The key information from the support and query branches is aggregated by the devised asymmetric-batched cross-attention.

\begin{figure}[!t]
	\centering
	\includegraphics[width=0.9\linewidth]{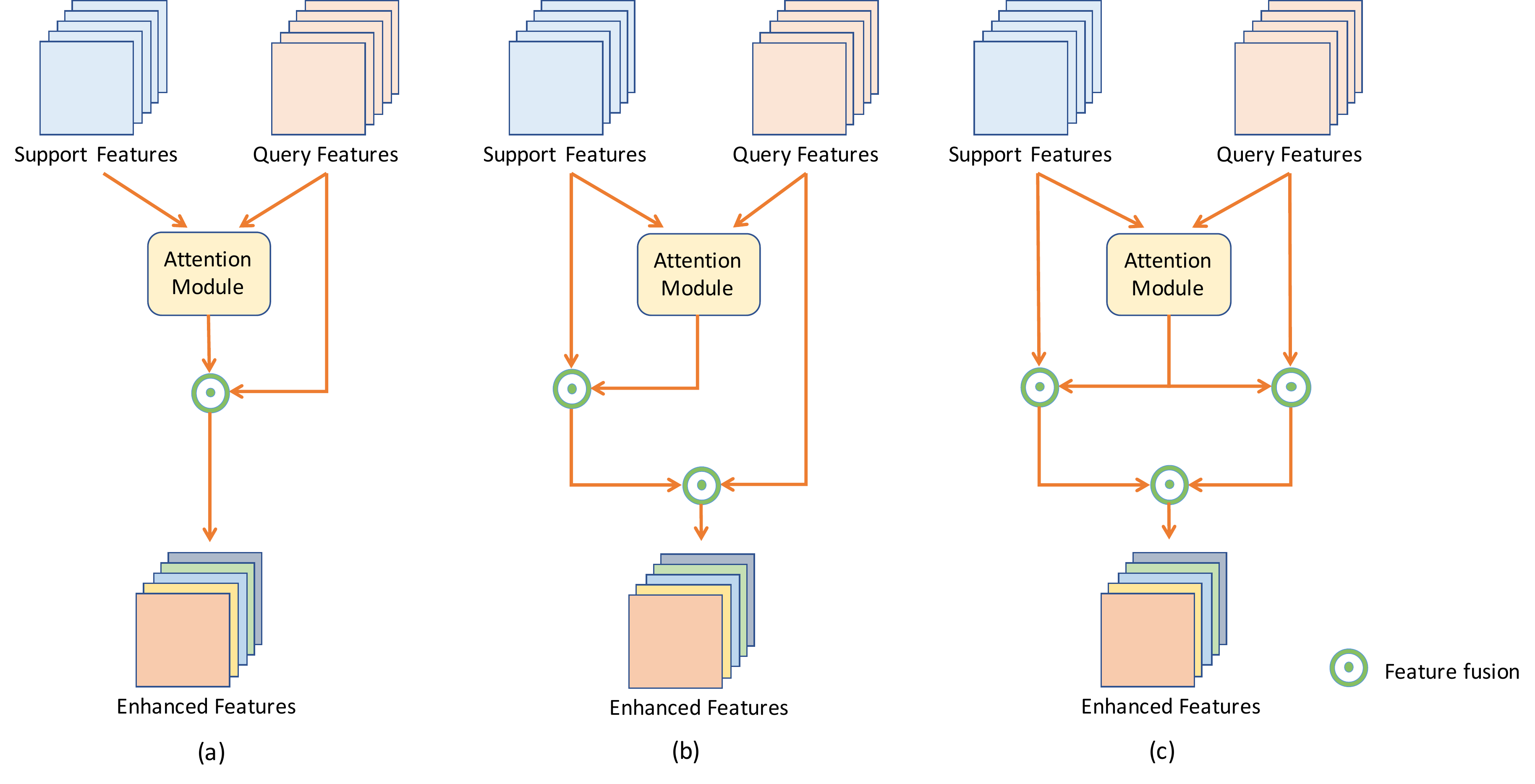}
	\caption{Structures of three different types of support and query feature aggregation methods. The first type (a) enhances only the query features by the support features~\cite{kangFewShot2019}. The second type (b) and third type (c) not only utilize the support features to enhance the query features but also aggregate the support features guided by the query features. Their difference lies in that the second type sequentially enhances the support and query features~\cite{chenDualAwareness2021, zhangAccurate2021}, while the third type simultaneously enhances them and then completes the aggregation~\cite{hsiehOneShot2019a, chenAdaptive2021, zhangMetaDETR2021}.}
	\label{fig:relation}
\end{figure}

Fig.~\ref{fig:relation} shows the structures of three types of attention-based support and query feature aggregation methods. The first type in Fig.~\ref{fig:relation}(a) enhances only the query features by the support features~\cite{kangFewShot2019}. The second and third types (Fig.~\ref{fig:relation}(b) and (c)) not only utilize the support features to enhance the query features, but also aggregate the support features guided by the query features. Their difference lies in that the second type sequentially enhances the support and query features~\cite{chenDualAwareness2021, zhangAccurate2021}, while the third type simultaneously enhances them and then completes the aggregation~\cite{hsiehOneShot2019a, chenAdaptive2021, zhangMetaDETR2021}.

In addition to attention-based FSOD methods, some models explicitly mine the relationship between the support and query features.
DRL~\cite{liuDynamic2021} adopts a dynamic Graph Convolutional Network (GCN)~\cite{kipf2016semi} to construct the relationship not only between the support and query features but also between different support categories.
The adjacency matrix consists of the similarities between features, which are calculated by \textit{Pearson's correlation coefficient}.
Each node in the graph represents either a support image or an RoI feature of the query image.
During training, the classification probabilities are updated according to the relationship between nodes.
Han et al.~\cite{hanQuery2021} introduced a heterogeneous graph with three types of edges and several subgraphs for effective relation discovery among support and query features.
The three types of edges, i.e., the class-class edges, the proposal-class edges, and the proposal-proposal edges, make the model capable of learning various relations between and within support classes and novel objects.
The subgraphs include a query-agnostic inter-class subgraph and multiple class-specific intra-class subgraphs.
The former one models multi-class relationships, while each of the latter ones constructs relations between the class and the query object proposals for each novel class.
GCN-FSOD~\cite{han2022few} proposes the local correspondence RPN where more fine-grained visual information is exploited to search for latent objects.
Moreover, an attentive detection head that mines local and global consistency for classification while exploiting class-attentive features for regression is designed to further improve the performance.

\begin{figure}[!t]
  \includegraphics[width=0.9\linewidth]{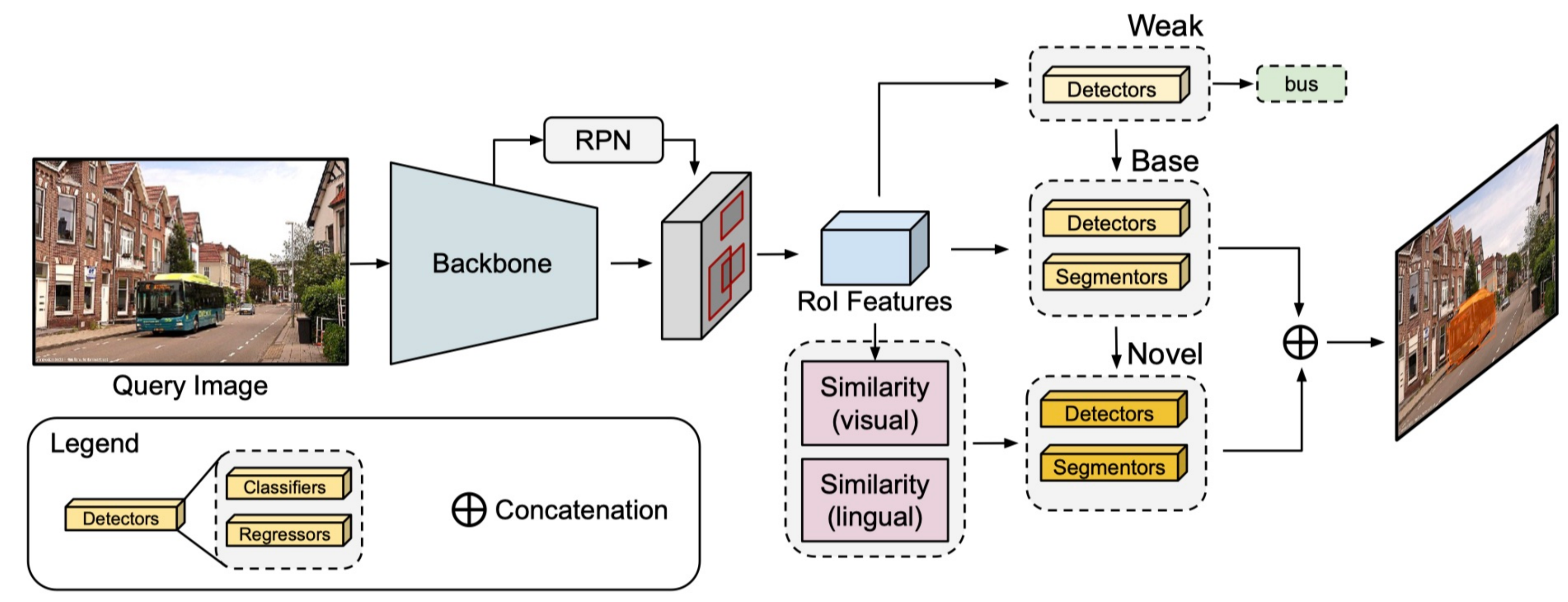}
  \caption{The framework of UniT~\cite{khandelwalUniT2021}. The key idea is to express each novel class as a weighted linear combination of base classes, which is realized via visual similarity and lingual similarity (pink box). During base training, the base detectors and segmentors are formed as the refinement on top of a weakly-supervised branch that is firstly trained by image-level annotations. In the novel fine-tuning stage, the detectors and segmentors are transferred from the base counterparts. Image courtesy of UniT~\cite{khandelwalUniT2021}.}
  \label{fig:unit}
\end{figure}

\subsubsection{\textbf{Learning from multi-tasks}}
Multitask learning~\cite{caruana1997multitask, zhang2017survey} intends to obtain the knowledge from several related tasks and generalize it to improve the performance on all the tasks.
Equ.~(\ref{eq_def}) is changed into:
\begin{equation}
\label{eq_multitask}
\begin{aligned}
	\theta = \underset{\theta}{\text{arg min}} \sum_{(x_i,c_i,t_i)\in D} & (\mathcal{L}^{cls}(\mathrm{CLS}(x_i),c_i;\theta^{B,S}_{T_1,...,T_n})\\
	&+\mathcal{L}^{loc}(\mathrm{LOC}(x_i),l_i;\theta^{B,S}_{T_1,...,T_n}))\\
	s.t. \quad c_i\in C_{novel}
\end{aligned}
\end{equation}
where $T_i$ ($i$=1, 2, ..., $n$) indicates the $i$-th task.

Segmentation is one of the fundamental computer vision tasks that aims to classify the pixels in images.
Several methods combine FSOD and segmentation into a framework.
Michaelis et al.~\cite{michaelis2018one} extended Mask R-CNN~\cite{he2017mask} with a siamese network to achieve one-shot object detection and instance segmentation.
The siamese network encodes the support and query image simultaneously, followed by a feature matching module.
Yan et al.~\cite{yanMeta2019} plugged a predictor-head remodeling network into Faster/Mask R-CNN~\cite{ren2015faster, he2017mask}.
The PRN share s the backbone with the original framework and generates class-attentive vectors via channel-wise soft-attention and average pooling.
The class-attentive vectors are finally fused with each RoI feature to improve the detection and segmentation performance.
Khandelwal et al.~\cite{khandelwalUniT2021} presented a semi-supervised framework UniT for few-shot (maybe zero-shot) object detection and segmentation, which is illustrated in Fig.~\ref{fig:unit}.
Different from the conventional semi-supervised methods, UniT utilizes the image-level annotated data instead of unlabeled data to train weak detectors.
The weak detectors are trained only during base training.
In the fine-tuning stage, the prior knowledge from the \textit{base set} is generalized into the \textit{novel set} based on additional visual and lingual similarities between the base and novel classes.

Furthermore, FSDetView~\cite{xiaoFewShot2020} proposes to tackle the problems of few-shot object detection and viewpoint estimation at the same time.
There are two feature encoders for the support and query inputs respectively.
For the FSOD task, the two encoders share the parameters, opposite to the few-shot viewpoint estimation where the parameters of two encoders are independent.
Each query feature is aggregated with all support features via a novel aggregation module.
A task-specific predictor finally generates the output.

Counting-DETR~\cite{nguyen2022few} tackles a new task called Few-Shot object Counting and Detection (FSCD), which utilizes abundant dot annotations with only a few exemplar bounding boxes to count and detect all boxes in the image. It regards objects as points, and the bounding boxes are directly regressed from the point features. The pseudo-labeling strategy is used to train the detector. Specifically, Counting-DETR is first trained to generate pseudo bounding boxes from the points, and then an uncertainty loss is proposed to refine the pseudo bounding boxes.

\begin{table}
\centering
\caption{A qualitative comparison among model-oriented FSOD methods.}
\label{tab:model-oriented}
\resizebox{\linewidth}{!}{
\begin{tabular}{ccccl}
\toprule
Category                                             & Model               & Year & Approach                          & Main Contribution                                                      \\
\bottomrule
\multirow{15}{*}{\makecell{Learning better \\embedding features}} & RepMet~\cite{karlinskyRepMet2019}              & 2019 & metric learning                   & first distance metric learning for FSOD                                \\
    & NP-RepMet~\cite{yangRestoring2020}           & 2020 & metric learning                   & exploiting negative instances                                             \\
    & DMNet~\cite{lu2022decoupled}               & 2022 & metric learning                   & parallel distance metric learning for multiple objects                         \\
    & Meta-RCNN~\cite{wuMetaRCNN2020}           & 2020 & class prototypes                  & first prototype-based FSOD method                                      \\
    & Han et al.~\cite{han2021meta}          & 2021 & class prototypes                  & spatial alignment between proposals and prototypes                     \\
    & FSOD$^{up}$~\cite{wuUniversalPrototype2021}                & 2021 & universal prototypes              & universal prototypes rather than class prototypes                      \\
    & Lee et al.~\cite{leeFewShot2021}          & 2021 & instance prototypes               & instances as prototypes                                                \\
    & CME~\cite{liMaxMargin2021}                 & 2021 & class-level feature refinement    & explicit feature space optimization via a new loss                     \\
    & SVD-Dictionary~\cite{wu2021generalized}           & 2021 & class-level feature refinement    & introducing SVD and dictionary learning                                  \\
    & FSCN~\cite{liFewShot2021a}           & 2021 & class-level feature refinement    & designing extra branch to reduce the number of false positives  \\
    & Wu et al.~\cite{wu2022learning}           & 2022 & class-level feature refinement    & suppressing imbalanced data distribution                                  \\
    & HiCLPL~\cite{zhang2022hierarchical}           & 2022 & class-level feature refinement    & hierarchical FSOD with hierarchical contrastive learning                       \\
    & Context-Transformer~\cite{yangContexttransformer2020} & 2020 & instance-level feature refinement & leveraging surrounding information of each box                        \\
    & FSCE~\cite{sunFSCE2021}                & 2021 & instance-level feature refinement & supervised contrastive learning for RoI feature refinement             \\
    & Kim et al.~\cite{kim2022few}                & 2022 & instance-level feature refinement & addressing IoU imbalance via fusing three detection heads             \\
    & MPSR~\cite{wuMultiscale2020}                & 2020 & instance-level feature refinement & dealing with scale variation problem in FSOD                              \\
\midrule
\multirow{15}{*}{\makecell{Learning support-query \\relations}}   & FSRW~\cite{kangFewShot2019}                & 2019 & meta-learning                     & first meta-learning FSOD framework                                     \\
    & Hsieh et al.~\cite{hsiehOneShot2019a}        & 2019 & attention mechanism               & proposing the co-attention and co-excitation framework                   \\
    & Attention-RPN~\cite{fanFewShot2020}       & 2020 & attention mechanism               & exploiting three different relations  a new FSOD dataset                  \\
    & DCNet~\cite{huDense2021}               & 2021 & attention mechanism               & pixel-level support-query match                                        \\
    & DAnA~\cite{chenDualAwareness2021}                & 2021 & attention mechanism               & suppressing background features  spatial misalignment                     \\
    & Zhang et al.~\cite{zhangAccurate2021}        & 2021 & attention mechanism               & additionally considering query guidance for support features              \\
    & KFSOD~\cite{zhang2022kernelized}               & 2022 & attention mechanism               & introducing kernelized functions for proposal generation                 \\
    & Park et al.~\cite{park2022hierarchical}         & 2022 & attention mechanism               & combining meta-learning with contrastive learning                        \\
    & Chen et al.~\cite{chenAdaptive2021}         & 2021 & attention mechanism               & emulating language translation process to extract proposal features      \\
    & Meta-DETR~\cite{zhangMetaDETR2021}           & 2022 & attention mechanism               & entire image-level detection framework based on Deformable DETR        \\
    & FS-DETR~\cite{bulat2022fs}           & 2022 & attention mechanism               & detecting novel objects without re-training via visual prompts       \\
    & FCT~\cite{han2022fct}           & 2022 & attention mechanism               & incorporating fully cross-transformer into Faster R-CNN        \\
    & DRL~\cite{liuDynamic2021}                 & 2021 & GCN~\cite{kipf2016semi}                               & additionally constructing class-to-class relations                        \\
    & QA-FewDet~\cite{hanQuery2021}          & 2021 & GCN~\cite{kipf2016semi}                               & class-to-class, class-to-proposal, and proposal-to-proposal relations  \\
    & GCN-FSOD~\cite{han2022few}          & 2022 & GCN~\cite{kipf2016semi}                               & fine-grained local correspondence between support and query features \\
\midrule
\multirow{5}{*}{\makecell{Learning from \\multi-tasks}}           & Michaelis et al.~\cite{michaelis2018one}    & 2018 & multi-task learning               & one-shot object detection and instance segmentation                    \\
    & Meta R-CNN~\cite{yanMeta2019}          & 2019 & multi-task learning               & few-shot object detection and image segmentation                       \\
    & UniT~\cite{khandelwalUniT2021}                & 2021 & multi-task learning               & any-shot object detection and image segmentation                       \\
    & FSDetView~\cite{xiaoFewShot2020}           & 2020 & multi-task learning               & few-shot object detection and viewpoint estimation                     \\
    & Counting-DETR~\cite{nguyen2022few}       & 2022 & multi-task learning               & proposing the few-shot object counting and detection task                    \\
\bottomrule
\end{tabular}}
\end{table}

\subsubsection{\textbf{Discussion and Conclusion}}
The majority of existing model-oriented FSOD methods concentrate on designing specific modules for the few-shot scenario.
A discriminative representation for each class is required to better distinguish confusable classes.
The clustering idea is widely adopted by these FSOD methods, learning either a good metric or a good class prototype.
Fully exploiting the information from the support set is also key to FSOD, whether through feature fusion or explicit relation learning.
Jointly training FSOD with some related tasks in a unified framework is another effective way, which usually plugs some special modules, but sometimes the framework becomes too complicated and many extra parameters are introduced.
The comparison among all model-oriented FSOD methods are shown in Table~\ref{tab:model-oriented}.

\subsection{Algorithm-oriented Methods}
Instead of devising novel modules, most methods of this type consider more about how to initialize the model parameters $\theta$ better or update $\theta$ faster, thus fewer parameter updates are required to reach the optimal target.
Some other methods try to extend the model's ability to detect novel classes and remember them. They are called incremental FSOD (iFSOD) methods, which are close to the open-world detection task.

\subsubsection{\textbf{Fine-tuning pre-trained parameters}}
The model trained on the \textit{base set} contains rich information of base classes, which is also helpful to initialize the novel detector.
Fine-tuning the pre-trained parameters on the \textit{novel set} is an effective way to obtain good performance.

\begin{figure}[!t]
  \includegraphics[width=0.9\linewidth]{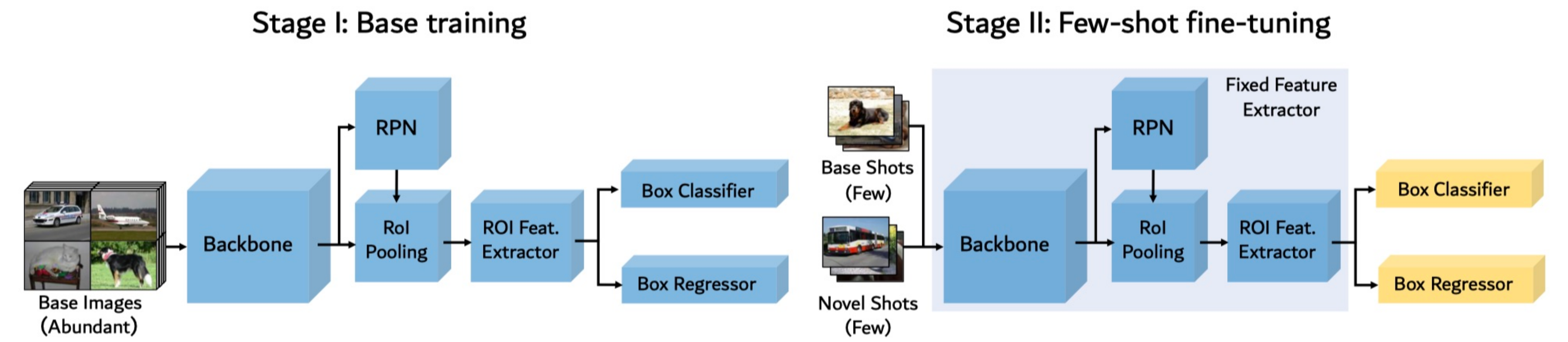}
  \caption{The TFA~\cite{wangFrustratingly2020} pipeline. There are two stages to train the model. In the base training stage, the model is trained with abundant labeled base images of base classes. In the few-shot fine-tuning stage, the model is provided with both base images and novel images, but there are only a few labeled instances for each class. The parameters except for that of the box classifier and regressor are frozen during the fine-tuning stage. Image courtesy of TFA~\cite{wangFrustratingly2020}.}
  \label{fig:tfa}
\end{figure}
LSTD~\cite{chenLSTD2018} is the first work that uses the fine-tuning strategy for the FSOD problem.
It combines Faster R-CNN~\cite{ren2015faster} with SSD~\cite{liu2016ssd} to design an effective framework for both regression and classification.
The regressor in SSD is class-agnostic, while that in Faster R-CNN is class-specific.
Thus, it is easy to take advantage of the pre-trained regression parameters in the target domain.
During fine-tuning, a novel regularization method with two modules is developed to handle the overfitting problem.
On the one hand, ground-truth boxes are used to restrain the disturbance of complex backgrounds.
On the other hand, the classifier pre-trained on base classes is considered as prior knowledge to regularize the training on novel classes.

TFA~\cite{wangFrustratingly2020} demonstrates that fine-tuning only the last layer of the pre-trained detector on novel classes can achieve a promising performance.
There are two stages in training TFA, one is base training, in which the model is trained on the \textit{base set} with abundant labeled images of base classes, and the other is the few-shot fine-tuning stage.
In the fine-tuning stage, the model is fed with both base images and novel images with only a few instances of each class, and all parameters are frozen except for the parameters of the box classifier and regressor.
Moreover, the cosine similarity is used to replace the original fully-connected layers to construct the classifier.
The pipeline is illustrated in Fig.~\ref{fig:tfa}.
Different from TFA~\cite{wangFrustratingly2020}, in the fine-tuning stage, Yang et al.~\cite{yang2022efficient} proposed to unfreeze RPN and only novel class parameters of the classifier.
Moreover, the novel class parameters are initialized via the proposed knowledge inheritance to reduce few-shot optimization steps. An adaptive length re-scaling strategy was devised to ensure the predicted novel weights have similar lengths to that of the base ones.

Retentive R-CNN~\cite{fanGeneralized2021} aims to keep the knowledge learned from the \textit{base set}, thus freezes the parameters of RPN and RoI heads, and adds some modules for novel classes.
Concretely, RPN adds a new branch that only fine-tunes the last layer to predict if the proposal is an object (which we call the objectness).
The output objectness is the maximum of the base branch and the fine-tuned one.
The RoI head also adopts a fine-tuning branch to generate predictions on both base and novel classes.
A consistency loss is proposed to ensure the performance on base classes.

\subsubsection{\textbf{Refining parameter update strategy}}
Different from the aforementioned fine-tuning methods, some FSOD methods intend to optimize the parameter update strategy.
Wang et al.~\cite{wangMetaLearning2019} classified the parameters in a CNN-based model into two types, i.e., category-agnostic and category-specific.
Concretely, category-agnostic parameters include the parameters of the convolutional features and RPN, while category-specific parameters include that of the classifiers and bounding box regressors.
The category-agnostic parameters are learned in the first training stage while frozen in the second stage.
A parameterized weight prediction meta-model is introduced for learning category-specific parameters from few samples.
Qiao et al.~\cite{qiaoDeFRCN2021} devised a modified Faster R-CNN framework DeFRCN from the perspective of gradient propagation.
First, due to the contradiction between the class-agnostic RPN and the class-specific RCNN, a gradient decoupled layer is designed to decouple them.
Second, as classification needs translation-invariant features while regression desires translation-covariant features, thus a novel metric-based score refinement module is used to decouple the two tasks.
CFA~\cite{guirguis2022cfa} proposes a constraint-based finetuning approach to alleviate catastrophic forgetting.
A new gradient update rule that imposes more constraints on the gradient search strategy is derived via a continual learning method, which allows for better knowledge exchange between base and novel classes.

\subsubsection{\textbf{Incremental FSOD}}
Incremental FSOD (iFSOD) demands the model to detect new classes with a few samples at any time without re-training or adaptation.
ONCE~\cite{perez-ruaIncremental2020} extends CenterNet~\cite{zhou2019objects} to iFSOD via decomposing it into class-agnostic and class-specific components.
The model is first trained on the \textit{base set} to obtain a class-agnostic feature extractor.
Then, a class code generator is trained still on the \textit{base set} with \textit{episodic training} strategy, during which the extractor trained in the first step is frozen.
Finally, the model is able to detect novel objects in a feed-forward manner without model re-training.
Sylph~\cite{yin2022sylph} improves ONCE~\cite{perez-ruaIncremental2020} in two ways. One is decoupling regression from classification through a detector pre-trained on abundant base classes. The other is a carefully designed class-conditional few-shot hypernetwork that improves classification accuracy.
Li et al.~\cite{liGeneralized2021} tackled iFSOD based on TFA~\cite{wangFrustratingly2020}.
To keep the performance on base classes, they disentangled the feature extraction into two branches for base and novel classes, respectively.
The catastrophic forgetting on novel classes is much more serious than that on base classes.
Consequently, the parameter search space is constrained to a tiny local region around the optimum of old classes detection via a novel stable moment matching algorithm.
In addition, RoI features of old classes are reserved, and a margin-based regularization loss is designed to correct the frequently misclassified old classes.
Incremental-DETR~\cite{dong2022incremental} proposes a two-stage fine-tuning strategy and utilizes self-supervised learning to retain the knowledge of base classes to learn better generalizable representations.
In the first stage, the model is first trained on the base images normally, and then only class-specific parameters are fine-tuned through a self-supervised way to make the model be generalized better on novel classes.
In the second stage, a knowledge distillation strategy is utilized to transfer knowledge from the base to the novel model using a few samples from each novel class.

\begin{table}
\centering
\caption{A qualitative comparison among algorithm-oriented FSOD methods.}
\resizebox{\linewidth}{!}{
\begin{tabular}{ccccl}
\toprule
Category   & Model    & Year & Approach     & Main Contribution           \\
\bottomrule
\multirow{4}{*}{Fine-tuning pre-trained parameters} & LSTD~\cite{chenLSTD2018}            & 2018 & knowledge regularization       & proposing the FSOD problem \& transfer learning framework                                \\
    & TFA~\cite{wangFrustratingly2020}             & 2020 & freezing partial parameters    & demonstrating the effectiveness of fine-tuning                        \\
    & Yang et al.~\cite{yang2022efficient}     & 2022 & freezing partial parameters    & proposing knowledge inheritance to boost parameter adaptation              \\
    & Retentive R-CNN~\cite{fanGeneralized2021} & 2021 & freezing partial parameters    & additional branch for novel class proposal generation \& classification         \\
\midrule
\multirow{3}{*}{Refining parameter update strategy} & Wang et al.~\cite{wangMetaLearning2019}     & 2019 & decoupling parameters    & decoupling RPN and RoI head parameters              \\
    & DeFRCN~\cite{qiaoDeFRCN2021}     & 2021 & decoupling parameters           & decoupling RPN and RoI head \& classification and regression parameters                  \\
    & CFA~\cite{guirguis2022cfa}     & 2022 & continuous learning      & imposing more constraint on the gradient search strategy        \\
\midrule
\multirow{4}{*}{Incremental FSOD}                   & ONCE~\cite{perez-ruaIncremental2020}            & 2020 & meta-learning                  & first iFSOD framework detecting novel classes without re-training                      \\
    & Sylph~\cite{yin2022sylph}           & 2022 & meta-learning                  & improving ONCE with hypernetwork-based framework \\
    & Li et al.~\cite{liGeneralized2021}       & 2021 & dynamic weights & extending TFA~\cite{wangFrustratingly2020} with individual and dynamic novel class weights                   \\
    & Incremental-DETR~\cite{dong2022incremental}       & 2022 & knowledge distillation & adapting DETR to iFSOD with a special two-stage strategy           \\

\bottomrule
\end{tabular}}
\end{table}

\subsubsection{\textbf{Discussion and Conclusion}}
Without specially designed modules, algorithm-oriented FSOD methods intend to obtain a good parameter initialization from prior knowledge or devise a faster parameter optimization algorithm.
Advantageously, not many extra parameters are introduced, but the weakness is the design of generic object detection frameworks may not suit the few-shot scenario and thus impact the detection performance.

iFSOD is a good direction that can be applied to the open-world detection task.
However, future methods should consider more about the ubiquitous domain shift problem in the few-shot scenario.
Moreover, the model may encounter the imbalance detection problem in real-world detection, which is quite different from the popular \textit{N}-way \textit{K}-shot problem.

\section{Performance Analysis and Comparison}\label{sec:analysis}
\begin{figure}[!t]
	\centering
	\includegraphics[width=0.95\linewidth]{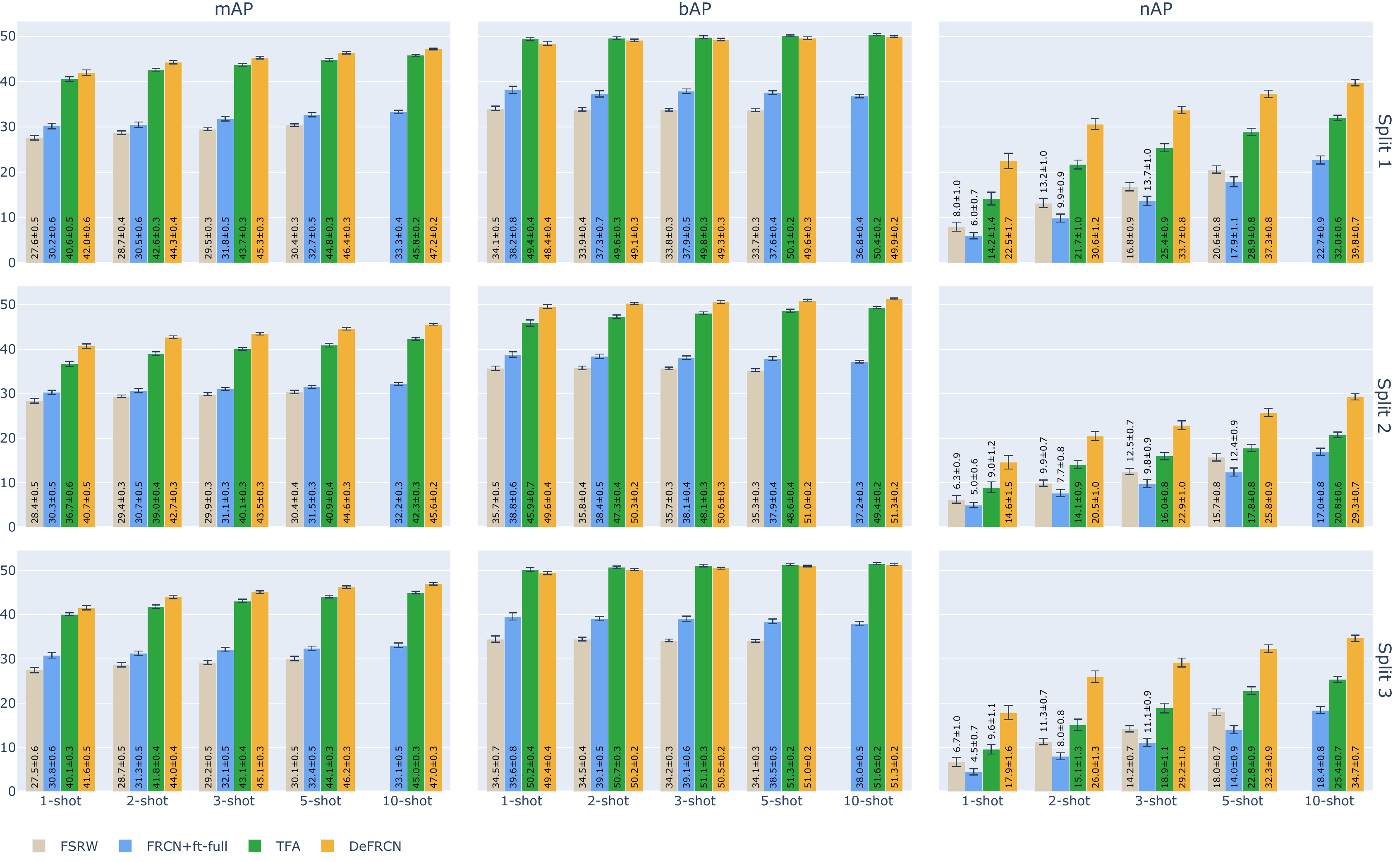}
	\caption{Generalized FSOD performance on PASCAL VOC~\cite{everingham2010pascal, kangFewShot2019}.
      The results of each metric are the averages and 95\% confidence interval computed over 30 random samples.
      All results are from~\cite{wangFrustratingly2020, qiaoDeFRCN2021}. mAP, bAP, and nAP indicate the mean AP, AP on the \textit{base set}, and AP on the \textit{novel set}, respectively.}
	\label{fig:bar_voc}
\end{figure}
\begin{figure}[!t]
	\centering
	\includegraphics[width=0.95\linewidth]{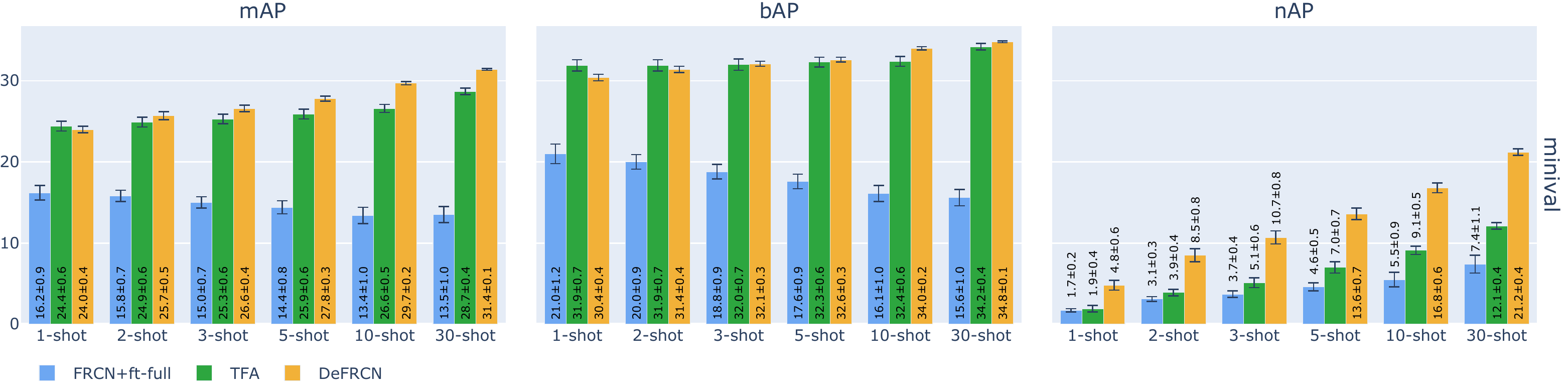}
	\caption{Generalized FSOD performance on MS-COCO \textit{minival} set~\cite{lin2014microsoft, kangFewShot2019}.
      The results of each metric are the averages and 95\% confidence interval computed over 30 random samples.
      All results are from~\cite{wangFrustratingly2020, qiaoDeFRCN2021}. mAP, bAP, and nAP indicate the mean AP, AP on the \textit{base set}, and AP on the \textit{novel set}, respectively.}
	\label{fig:bar_coco}
\end{figure}

Based on our taxonomy, we present a performance comparison of existing FSOD methods on the VOC novel sets and MS-COCO \textit{minival} set in Tab.~\ref{tab:voc} and Tab.~\ref{tab:coco}, respectively.
Additionally, we illustrate the generalized FSOD (G-FSOD) performance of several typical methods in Fig.~\ref{fig:bar_voc} and Fig.~\ref{fig:bar_coco}.
G-FSOD was proposed by Wang et al.~\cite{wangFrustratingly2020}, which demands the model to run on different random training samples multiple times and report the performance on both base and novel classes.
Therefore, it is able to check the robustness of each FSOD method.

\begin{table}[tbp]
  \footnotesize
  \centering
  \caption{FSOD performance on PASCAL VOC novel sets (AP50)~\cite{kangFewShot2019}.
  {\color{red}RED}/{\color{blue}BLUE} indicates state-of-the-art/second-best performance.
  * indicates mean multiple-run result (30 random seeds).}
  \resizebox{\linewidth}{!}{
    \begin{tabular}{rrcccccccccccccccc}
    \toprule
    \multicolumn{2}{c|}{\multirow{2}[2]{*}{Method}} & \multicolumn{1}{c|}{\multirow{2}[2]{*}{Framework}} & \multicolumn{5}{c|}{Novel Split 1}    & \multicolumn{5}{c|}{Novel Split 2}    & \multicolumn{5}{c}{Novel Split 3} \\
    \multicolumn{2}{c|}{} & \multicolumn{1}{c|}{} & 1-shot & 2-shot & 3-shot & 5-shot & \multicolumn{1}{c|}{10-shot} & 1-shot & 2-shot & 3-shot & 5-shot & \multicolumn{1}{c|}{10-shot} & 1-shot & 2-shot & 3-shot & 5-shot & 10-shot \\
    \midrule
    \multicolumn{1}{c}{\multirow{6}[1]{*}{\rotatebox{90}{\makecell{Data\\-Oriented\\ Methods}}}} & \multicolumn{1}{l|}{TIP~\cite{liTransformation2021}} & \multicolumn{1}{c|}{Faster R-CNN} & 27.7  & 36.5  & 43.3  & 50.2  & \multicolumn{1}{c|}{59.6} & 22.7  & 30.1  & 33.8  & 40.9  & \multicolumn{1}{c|}{46.9} & 21.7  & 30.6  & 38.1  & 44.5  & 50.9 \\
          & \multicolumn{1}{l|}{Halluc~\cite{zhangHallucination2021}} & \multicolumn{1}{c|}{Faster R-CNN} & 47.0    & 44.9  & 46.5  & 54.7  & \multicolumn{1}{c|}{54.7} & 26.3  & 31.8  & 37.4  & 37.4  & \multicolumn{1}{c|}{41.2} & 40.4  & 42.1  & 43.3  & 51.4  & 49.6 \\
          & \multicolumn{1}{l|}{SRR-FSD~\cite{zhuSemantic2021}} & \multicolumn{1}{c|}{Faster R-CNN} & 47.8  & 50.5  & 51.3  & 55.2  & \multicolumn{1}{c|}{56.8} & 32.5  & 35.3  & 39.1  & 40.8  & \multicolumn{1}{c|}{43.8} & 40.1  & 41.5  & 44.3  & 46.9  & 46.4 \\
          & \multicolumn{1}{l|}{FADI~\cite{cao2021few}} & \multicolumn{1}{c|}{Faster R-CNN} & 50.3 & 54.8 & 54.2  & 59.3  & \multicolumn{1}{c|}{63.2} & 30.6  & 35.0    & 40.3  & 42.8  & \multicolumn{1}{c|}{48.0} & 45.7  & 49.7  & 49.1  & 55.0 & 59.6 \\
          & \multicolumn{1}{l|}{MINI~\cite{cao2022mini}} & \multicolumn{1}{c|}{Faster R-CNN} & \textbf{\color{red}72.0} & \textbf{\color{red}74.2} & \textbf{\color{red}72.4}  & \textbf{\color{red}74.7}  & \multicolumn{1}{c|}{\textbf{\color{red}76.2}} & \textbf{\color{red}51.8}  & \textbf{\color{red}53.6}    & \textbf{\color{red}62.3}  & \textbf{\color{red}62.1}  & \multicolumn{1}{c|}{\textbf{\color{red}63.7}} & \textbf{\color{red}65.0} & \textbf{\color{red}66.5}  & \textbf{\color{red}64.0}  & \textbf{\color{red}66.9} & \textbf{\color{red}70.4} \\
          & \multicolumn{1}{l|}{Kaul et al.~\cite{kaul2022label}} & \multicolumn{1}{c|}{Faster R-CNN} & 54.5 & 53.2 & 58.8  & 63.2  & \multicolumn{1}{c|}{65.7} & 32.8  & 29.2    & 50.7  & 49.8  & \multicolumn{1}{c|}{50.6} & 48.4  & 52.7  & 55.0  & 59.6 & 59.6 \\
    \midrule
    \multicolumn{1}{c}{\multirow{21}[1]{*}{\rotatebox{90}{\makecell{Model-Oriented\\ Methods}}}} & \multicolumn{1}{l|}{RepMet~\cite{karlinskyRepMet2019}} & \multicolumn{1}{c|}{Faster R-CNN} & 26.1  & 32.9  & 34.4  & 38.6  & \multicolumn{1}{c|}{41.3} & 17.2  & 22.1  & 23.4  & 28.3  & \multicolumn{1}{c|}{35.8} & 27.5  & 31.1  & 31.5  & 34.4  & 37.2 \\
          & \multicolumn{1}{l|}{NP-RepMet~\cite{yangRestoring2020}} & \multicolumn{1}{c|}{Faster R-CNN} & 37.8  & 40.3  & 41.7  & 47.3  & \multicolumn{1}{c|}{49.4} & \textbf{\color{blue}41.6} & 43.0 & 43.4  & 47.4  & \multicolumn{1}{c|}{49.1} & 33.3  & 38.0    & 39.8  & 41.5  & 44.8 \\
          & \multicolumn{1}{l|}{DMNet~\cite{lu2022decoupled}} & \multicolumn{1}{c|}{FPN} & 39.0  & 48.9  & 50.7  & 58.6  & \multicolumn{1}{c|}{62.5} & 31.2 & 32.4 & 40.3  & 47.6  & \multicolumn{1}{c|}{52.0} & 41.7  & 41.8    & 42.7  & 50.3  & 52.1 \\
          & \multicolumn{1}{l|}{FSOD$^{up}$~\cite{wuUniversalPrototype2021}} & \multicolumn{1}{c|}{Faster R-CNN} & 43.8  & 47.8  & 50.3  & 55.4  & \multicolumn{1}{c|}{61.7} & 31.2  & 30.5  & 41.2  & 42.2  & \multicolumn{1}{c|}{48.3} & 35.5  & 39.7  & 43.9  & 50.6  & 53.5 \\
          & \multicolumn{1}{l|}{Lee et al.~\cite{leeFewShot2021}} & \multicolumn{1}{c|}{Faster R-CNN} & 31.1  & 36.1  & 39.2  & 50.7  & \multicolumn{1}{c|}{59.4} & 22.9  & 29.4  & 32.1  & 35.4  & \multicolumn{1}{c|}{42.7} & 24.3  & 28.6  & 35.0    & 50.0    & 53.6 \\
          & \multicolumn{1}{l|}{CME~\cite{liMaxMargin2021}} & \multicolumn{1}{c|}{Faster R-CNN} & 41.5  & 47.5  & 50.4  & 58.2  & \multicolumn{1}{c|}{60.9} & 27.2  & 30.2  & 41.4  & 42.5  & \multicolumn{1}{c|}{46.8} & 34.3  & 39.6  & 45.1  & 48.3  & 51.5 \\
          & \multicolumn{1}{l|}{SVD-Dictionary~\cite{wu2021generalized}} & \multicolumn{1}{c|}{Faster R-CNN} & 46.1  & 43.5  & 48.9  & 60.0    & \multicolumn{1}{c|}{61.7} & 25.6  & 29.9  & 44.8  & 47.5  & \multicolumn{1}{c|}{48.2} & 39.5  & 45.4  & 48.9  & 53.9  & 56.9 \\
          & \multicolumn{1}{l|}{FSCN~\cite{liFewShot2021a}} & \multicolumn{1}{c|}{Faster R-CNN} & 40.7  & 45.1  & 46.5  & 57.4  & \multicolumn{1}{c|}{62.4} & 27.3  & 31.4  & 40.8  & 42.7  & \multicolumn{1}{c|}{46.3} & 31.2  & 36.4  & 43.7  & 50.1  & 55.6 \\
          & \multicolumn{1}{l|}{Wu et al.~\cite{wu2022learning}} & \multicolumn{1}{c|}{Faster R-CNN} & 29.9  & 42.2  & 46.1  & 53.3  & \multicolumn{1}{c|}{59.6} & 25.5  & 28.4  & 34.9  & 39.8  & \multicolumn{1}{c|}{48.6} & 28.7  & 33.6  & 40.6  & 46.8  & 51.0 \\
          & \multicolumn{1}{l|}{FSCE~\cite{sunFSCE2021}} & \multicolumn{1}{c|}{Faster R-CNN} & 44.2  & 43.8  & 51.4  & 61.9  & \multicolumn{1}{c|}{63.4} & 27.3  & 29.5  & 43.5  & 44.2  & \multicolumn{1}{c|}{50.2} & 37.2  & 41.9  & 47.5  & 54.6  & 58.5 \\
          & \multicolumn{1}{l|}{Kim et al.~\cite{kim2022few}} & \multicolumn{1}{c|}{Faster R-CNN} & 39.2  & 49.2  & 56.4  & 57.4  & \multicolumn{1}{c|}{61.6} & 28.7  & 31.3  & 36.9  & 37.4  & \multicolumn{1}{c|}{44.3} & 37.4  & 44.3  & 48.6  & 51.8  & 56.0 \\
          & \multicolumn{1}{l|}{MPSR~\cite{wuMultiscale2020}} & \multicolumn{1}{c|}{Faster R-CNN} & 41.7  &    -  & 51.4  & 55.2  & \multicolumn{1}{c|}{61.8} & 24.4  &   -   & 39.2  & 39.9  & \multicolumn{1}{c|}{47.8} & 35.6  &   -   & 42.3  & 48.0    & 49.7 \\
          & \multicolumn{1}{l|}{FSRW~\cite{kangFewShot2019}} & \multicolumn{1}{c|}{YOLO-V2} & 14.8  & 15.5  & 26.7  & 33.9  & \multicolumn{1}{c|}{47.2} & 15.7  & 15.3  & 22.7  & 30.1  & \multicolumn{1}{c|}{40.5} & 21.3  & 25.6  & 28.4  & 42.8  & 45.9 \\
          & \multicolumn{1}{l|}{Zhang et al.~\cite{zhangAccurate2021}} & \multicolumn{1}{c|}{Faster R-CNN} & 48.6  & 51.1  & 52.0    & 53.7  & \multicolumn{1}{c|}{54.3} & \textbf{\color{blue}41.6} & 45.4 & 45.8  & 46.3  & \multicolumn{1}{c|}{48.0} & 46.1 & 51.7 & 52.6 & 54.1  & 55.0 \\
          & \multicolumn{1}{l|}{KFSOD~\cite{zhang2022kernelized}} & \multicolumn{1}{c|}{Faster R-CNN} & 44.6  & -  & 54.4    & 60.9  & \multicolumn{1}{c|}{65.8} & 37.8 & - & 43.1  & 48.1  & \multicolumn{1}{c|}{50.4} & 34.8 & - & 44.1 & 52.7  & 53.9 \\
          & \multicolumn{1}{l|}{Meta-DETR~\cite{zhangMetaDETR2021}} & \multicolumn{1}{c|}{Deformable DETR} & 40.6  & 51.4  & 58.0 & 59.2  & \multicolumn{1}{c|}{63.6} & 37.0 & 36.6  & 43.7  & 49.1 & \multicolumn{1}{c|}{\textbf{\color{blue}54.6}} & 41.6  & 45.9  & 52.7 & 58.9 & 60.6 \\
          & \multicolumn{1}{l|}{FS-DETR~\cite{bulat2022fs}} & \multicolumn{1}{c|}{Conditional DETR} & 45.0  & 48.5  & 51.5    & 52.7  & \multicolumn{1}{c|}{56.1} & 37.3 & 41.3 & 43.4  & 46.6  & \multicolumn{1}{c|}{49.0} & 43.8 & 47.1 & 50.6 & 52.1  & 56.9 \\
          & \multicolumn{1}{l|}{FCT~\cite{han2022fct}} & \multicolumn{1}{c|}{Faster R-CNN} & 49.9  & 57.1  & 57.9    & 63.2  & \multicolumn{1}{c|}{67.1} & 27.6 & 34.5 & 43.7  & 49.2  & \multicolumn{1}{c|}{51.2} & 39.5 & 54.7 & 52.3 & 57.0  & 58.7 \\
          & \multicolumn{1}{l|}{QA-FewDet~\cite{hanQuery2021}} & \multicolumn{1}{c|}{Faster R-CNN} & 42.4  & 51.9  & 55.7  & 62.6 & \multicolumn{1}{c|}{63.4} & 25.9  & 37.8  & 46.6 & 48.9  & \multicolumn{1}{c|}{51.1} & 35.2  & 42.9  & 47.8  & 54.8  & 53.5 \\
          & \multicolumn{1}{l|}{DCNet~\cite{huDense2021}} & \multicolumn{1}{c|}{Faster R-CNN} & 33.9  & 37.4  & 43.7  & 51.1  & \multicolumn{1}{c|}{59.6} & 23.2  & 24.8  & 30.6  & 36.7  & \multicolumn{1}{c|}{46.6} & 32.3  & 34.9  & 39.7  & 42.6  & 50.7 \\
          & \multicolumn{1}{l|}{Meta R-CNN~\cite{yanMeta2019}} & \multicolumn{1}{c|}{Faster R-CNN} & 19.9  & 25.5  & 35.0    & 45.7  & \multicolumn{1}{c|}{51.5} & 10.4  & 19.4  & 29.6  & 34.8  & \multicolumn{1}{c|}{45.4} & 14.3  & 18.2  & 27.5  & 41.2  & 48.1 \\
    \midrule
    \multicolumn{1}{c}{\multirow{6}[2]{*}{\rotatebox{90}{\makecell{Algorithm\\-Oriented\\ Methods}}}} & \multicolumn{1}{l|}{TFA~\cite{wangFrustratingly2020}} & \multicolumn{1}{c|}{Faster R-CNN} & 39.8  & 36.1  & 44.7  & 55.7  & \multicolumn{1}{c|}{56.0} & 23.5  & 26.9  & 34.1  & 35.1  & \multicolumn{1}{c|}{39.1} & 30.8  & 34.8  & 42.8  & 49.5  & 49.8 \\
    	  & \multicolumn{1}{l|}{Yang et al.~\cite{yang2022efficient}} & \multicolumn{1}{c|}{Faster R-CNN} & 41.5  & 48.7  & 49.2  & 57.6  & \multicolumn{1}{c|}{60.9} & 26.0  & 34.4  & 39.6  & 40.1    & \multicolumn{1}{c|}{46.4} & 37.1  & 44.4  & 44.6    & 53.2  & 54.9 \\
          & \multicolumn{1}{l|}{Retentive R-CNN~\cite{fanGeneralized2021}} & \multicolumn{1}{c|}{Faster R-CNN} & 42.4  & 45.8  & 45.9  & 53.7  & \multicolumn{1}{c|}{56.1} & 21.7  & 27.8  & 35.2  & 37.0    & \multicolumn{1}{c|}{40.3} & 30.2  & 37.6  & 43.0    & 49.7  & 50.1 \\
          & \multicolumn{1}{l|}{MetaDet~\cite{wangMetaLearning2019}} & \multicolumn{1}{c|}{Faster R-CNN} & 18.9  & 20.6  & 30.2  & 36.8  & \multicolumn{1}{c|}{49.6} & 21.8  & 23.1  & 27.8  & 31.7  & \multicolumn{1}{c|}{43.0} & 20.6  & 23.9  & 29.4  & 43.9  & 44.1 \\
          & \multicolumn{1}{l|}{DeFRCN~\cite{qiaoDeFRCN2021}} & \multicolumn{1}{c|}{Faster R-CNN} & 53.6 & 57.5 & 61.5 & 64.1 & \multicolumn{1}{c|}{60.8} & 30.1  & 38.1  & 47.0 & 53.3 & \multicolumn{1}{c|}{47.9} & 48.4 & 50.9 & 52.3  & 54.9  & 57.4 \\
          & \multicolumn{1}{l|}{CFA~\cite{guirguis2022cfa}} & \multicolumn{1}{c|}{Faster R-CNN} & \textbf{\color{blue}59.0}  & \textbf{\color{blue}63.5}  & \textbf{\color{blue}66.4}  & \textbf{\color{blue}68.4}  & \multicolumn{1}{c|}{\textbf{\color{blue}68.3}} & 37.0  & \textbf{\color{blue}45.8}  & \textbf{\color{blue}50.0}  & \textbf{\color{blue}54.2}  & \multicolumn{1}{c|}{52.5} & \textbf{\color{blue}54.8}  & \textbf{\color{blue}58.5}  & \textbf{\color{blue}56.5}  & \textbf{\color{blue}61.3}  & \textbf{\color{blue}63.5}\\
    \bottomrule
    \multicolumn{1}{c}{\multirow{6}[2]{*}{}} & \multicolumn{1}{l|}{Faster R-CNN-ft*~\cite{yanMeta2019}} & \multicolumn{1}{c|}{Faster R-CNN} & 9.9  & 15.6  & 21.6  & 28.0  & \multicolumn{1}{c|}{52.0} & 9.4  & 13.8  & 17.4  & 21.9  & \multicolumn{1}{c|}{39.7} & 8.1  & 13.9  & 19.0  & 23.9  & 44.6 \\
          & \multicolumn{1}{l|}{Lee et al.*~\cite{leeFewShot2021}} & \multicolumn{1}{c|}{Faster R-CNN} & 24.3  & 36.5  & 44.9  & 52.0  & \multicolumn{1}{c|}{59.2} & 20.5  & 27.5  & 33.1  & 40.9  & \multicolumn{1}{c|}{47.1} & 22.4  & 33.0  & 37.8  & 43.9  & 51.5 \\
          & \multicolumn{1}{l|}{FCT*~\cite{han2022fct}} & \multicolumn{1}{c|}{Faster R-CNN} & 38.5  & 49.6  & 53.5    & 59.8  & \multicolumn{1}{c|}{64.3} & 25.9 & 34.2 & 40.1  & 44.9  & \multicolumn{1}{c|}{47.4} & 34.7 & 43.9 & 49.3 & 53.1  & 56.3 \\
          & \multicolumn{1}{l|}{TFA*~\cite{wangFrustratingly2020}} & \multicolumn{1}{c|}{Faster R-CNN} & 25.3  & 36.4  & 42.1  & 47.9  & \multicolumn{1}{c|}{52.8} & 18.3  & 27.5  & 30.9  & 34.1    & \multicolumn{1}{c|}{39.5} & 17.9  & 27.2  & 34.3    & 40.8  & 45.6 \\
          & \multicolumn{1}{l|}{FSDetView*~\cite{xiaoFewShot2020}} & \multicolumn{1}{c|}{Faster R-CNN} & 24.2  & 35.3  & 42.2  & 49.1  & \multicolumn{1}{c|}{57.4} & 21.6  & 24.6  & 31.9  & 37.0    & \multicolumn{1}{c|}{45.7} & 21.2  & 30.0  & 37.2    & 43.8  & 49.6 \\
          & \multicolumn{1}{l|}{DeFRCN*~\cite{qiaoDeFRCN2021}} & \multicolumn{1}{c|}{Faster R-CNN} & 40.2 & 53.6 & 58.2 & 63.6 & \multicolumn{1}{c|}{66.5} & 29.5  & 39.7  & 43.4 & 48.1 & \multicolumn{1}{c|}{52.8} & 35.0 & 38.3 & 52.9  & 57.7  & 60.8 \\
    \bottomrule
    \end{tabular}}%
  \label{tab:voc}%
\end{table}

\begin{figure}[!t]
	\centering
	\includegraphics[width=0.75\linewidth]{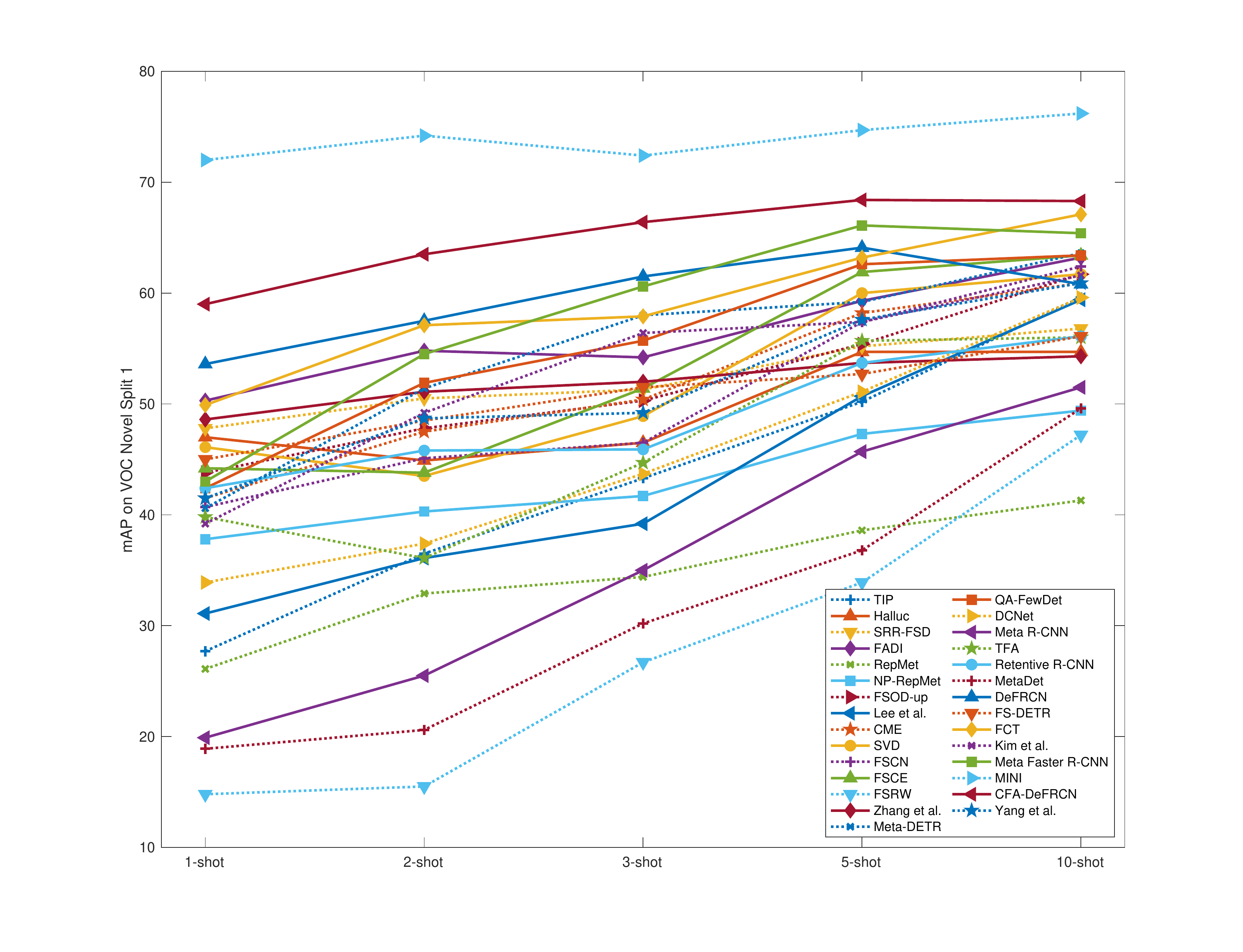}
	\caption{Performance (AP50) comparison of existing FSOD methods on PASCAL VOC split 1~\cite{kangFewShot2019}.}
	\label{fig:voc_line}
\end{figure}

As we can see in Tab.~\ref{tab:voc}, the performance is highly positively correlated with the number of shots.
For instance, DeFRCN~\cite{qiaoDeFRCN2021} obtains 53.6 AP of 1-shot detection while 60.8 AP of 10-shot detection in novel split 1, where there is a significant gap of 7.2 AP.
Fig.~\ref{fig:voc_line} provides a more obvious increasing trend when the number of shots increases.
Although MINI~\cite{cao2022mini} mines many implicit novel objects in the base images and achieves high performance in all novel splits, there is still a non-negligible performance gap between the 1-shot and 10-shot settings.
Based on the performance analysis of FSOD methods with different number of shots, we can see that object detection with very few samples is much more challenging than that with sufficient data.

The bottom of Tab.~\ref{tab:voc} reports the average performance over multiple experiments with different random seeds of each method.
It is shown that for the same method, the performance of multiple-run is much worse than that of single-run when \textit{K} is small.
For instance, when \textit{K} is 1, DeFRCN~\cite{qiaoDeFRCN2021} obtains 40.2 AP of multiple-run while 53.6 AP of single-run in novel split 1, where the gap is over 10 AP.
Besides, Fig.~\ref{fig:bar_voc} and Fig.~\ref{fig:bar_coco} show that the 95\% confidence interval is larger when \textit{K} is smaller.
This indicates that the sample variance seriously influences existing FSOD methods' performance.
Moreover, there is still a far cry from the performance on the \textit{novel set} to that on the \textit{base set}.

\begin{table}[tbp]
  \footnotesize
  \centering
  \caption{FSOD performance on MS-COCO \textit{minival} novel set~\cite{kangFewShot2019}.
  {\color{red}RED}/{\color{blue}BLUE} indicates state-of-the-art/second-best performance.
  AP$_{50}$ and AP$_{75}$ indicate the performance under the IoU threshold of 0.5 and 0.75, respectively.
  AP$_{S}$, AP$_{M}$, and AP$_{L}$ separately represent the performance on small, medium, and large objects.
  * indicates mean multiple-run result (30 random seeds).}
  \resizebox{\linewidth}{!}{
    \begin{tabular}{cl|c|cc|cc|cc|cc|cc|cc}
    \toprule
    \multicolumn{2}{c|}{\multirow{2}{*}{Method}} & \multirow{2}{*}{Framework} & \multicolumn{2}{c|}{nAP} & \multicolumn{2}{c|}{nAP$_{50}$} & \multicolumn{2}{c|}{nAP$_{75}$} & \multicolumn{2}{c|}{nAP$_{S}$} & \multicolumn{2}{c|}{nAP$_{M}$} & \multicolumn{2}{c}{nAP$_{L}$} \\
    \multicolumn{2}{c|}{} &       & 10-shot & 30-shot & 10-shot & 30-shot & 10-shot & 30-shot & 10-shot & 30-shot & 10-shot & 30-shot & 10-shot & 30-shot \\
    \midrule
    \multicolumn{1}{c}{\multirow{5}{*}{\rotatebox{90}{\makecell{Data-\\Oriented \\Methods}}}} & TIP~\cite{liTransformation2021}   & Faster R-CNN & 16.3  & 18.3  & 33.2  & 35.9  & 14.1  & 16.9  & 5.4   & 6.0     & \textbf{\color{blue}17.5}  & \textbf{\color{blue}19.3}  & 25.8  & \textbf{\color{blue}29.2} \\
          & SRR-FSD~\cite{zhuSemantic2021}   & Faster R-CNN & 11.3  & 14.7  & 23    & 29.2  & 9.8   & 13.5  &   -   &   -   &   -   &  -     &   -   & - \\
          & FADI~\cite{cao2021few}    & Faster R-CNN & 12.2  & 16.1  &   -   &-       & 11.9  & 15.8  &   -   &   -   &   -   &   -   &  -     &  -\\
          & MINI~\cite{cao2022few}    & Faster R-CNN & \textbf{\color{blue}21.8}  & \textbf{\color{red}27.3}  &   \textbf{\color{blue}38.0}   &  \textbf{\color{blue}44.9}     & \textbf{\color{blue}21.5}  & \textbf{\color{red}28.5}  &   -   &   -   &   -   &   -   &  -     &  -\\
          & Kaul et al.~\cite{kaul2022label}    & Faster R-CNN & 17.6  & 24.5  &   30.9   & 41.1       & 17.8  & \textbf{\color{blue}25.0}  &   -   &   -   &   -   &   -   &  -     &  -\\
    \midrule
    \multicolumn{1}{c}{\multirow{22}{*}{\rotatebox{90}{\makecell{Model-Oriented\\ Methods}}}} &
    	  DMNet~\cite{lu2022decoupled}  & FPN & 10.0   & 17.1  & 17.4  & 29.7  & 10.4     & 17.7  & 3.4   & 4.8   & 8.3     & 14.7  & 16.1  & 26.5 \\
    	  & Meta-RCNN~\cite{wuMetaRCNN2020}  & Faster R-CNN & 9.5   & 12.8  & 19.9  & 27.3  & 7.0     & 11.4  & 2.2   & 2.6   & 9.0     & 12.4  & 14.6  & 19.1 \\
          & FSOD-up~\cite{wuUniversalPrototype2021}  & Faster R-CNN & 11.0    & 15.6  &  -     &   -   & 10.7  & 15.7  & 4.5   & 4.7   & 11.2  & 15.1  & 17.3  & 25.1 \\
          & Lee et al.~\cite{leeFewShot2021}   & Faster R-CNN & 13.0    & 15.3  & 24.7  & 29.3  & 12.1  & 14.5  &   -   &   -   &  -     &   -   &   -   &  -\\
          & CME~\cite{liMaxMargin2021}    & Faster R-CNN & 15.1  & 16.9  & 24.6  & 28.0    & 16.4  & 17.8  & 4.6   & 4.6   & 16.6  & 18.0    & \textbf{\color{blue}26.0}    & \textbf{\color{blue}29.2} \\
          & SVD-Dictionary~\cite{wu2021generalized}   & Faster R-CNN & 12.0    & 16.0    &  -    &-       & 10.4  & 15.3  & 4.2   & 6.0    & 12.1  & 16.8  & 18.9  & 24.9 \\
          & FSCN~\cite{liFewShot2021a}   & Faster R-CNN & 11.3  & 15.1  & 20.3  & 29.4  &   -   &   -   &   -   &   -   &   -   &  -    &   -   & - \\
          & Wu et al.~\cite{wu2022learning}   & Faster R-CNN & 13.3  & 15.8  & 27.8  & 30.4  &   11.2   &   12.8   &   2.2   &   3.2   &   14.5   &  16.1    &   20.4   & 25.1 \\
          & FSCE~\cite{sunFSCE2021}   & Faster R-CNN & 11.9  & 16.4  &       &       & 10.5  & 16.2  &   -   &   -   &   -   &   -   &  -    & - \\
          & MPSR~\cite{wuMultiscale2020} & Faster R-CNN & 9.8   & 14.1  & 17.9  & 25.4  & 9.7   & 14.2  & 3.3   & 4.0     & 9.2   & 12.9  & 16.1  & 23.0 \\
          & FSRW~\cite{kangFewShot2019}  & YOLO-V2 & 5.6   & 9.1   & 12.3  & 19.0    & 4.6   & 7.6   & 0.9   & 0.8   & 3.5   & 4.9   & 10.5  & 16.8 \\
          & DAnA~\cite{chenDualAwareness2021}  & Faster R-CNN & 18.6  & 21.6  &   -   &-       & 17.2  & 20.3  &   -   &   -   &   -   &   -   &  -    &  -\\
          & Zhang et al.~\cite{zhangAccurate2021} & Faster R-CNN & 13.9  &   -   & 29.5  &   -   & 11.7  &   -   & \textbf{\color{blue}7.6}   &   -   & 15.2  &  -    & 19.0    &  -\\
          & KFSOD~\cite{zhang2022kernelized} & Faster R-CNN & 18.5  &   -   & 26.3  &   -   & 18.7  &   -   & -   &   -   & -  &  -    & -    &  -\\
          & Park et al.~\cite{park2022hierarchical} & Faster R-CNN & \textbf{\color{red}22.4} &   \textbf{\color{blue}25.0}   & \textbf{\color{red}40.7}  &   \textbf{\color{red}45.2}   & \textbf{\color{red}22.0}  &   24.6   & \textbf{\color{red}8.2}   &   \textbf{\color{red}7.9}   & \textbf{\color{red}19.2}  &  \textbf{\color{red}22.4}    & \textbf{\color{red}34.3}    &  \textbf{\color{red}37.7}\\
          & Meta-DETR~\cite{zhangMetaDETR2021} & Deformable DETR & 19.0    & 22.2  & 30.5  & 35.0    & 19.7  & 22.8  &   -   &    -  &  -    &   -   &   -   &  -\\
          & FS-DETR~\cite{bulat2022fs} & Conditional DETR & 11.1    & -  & 21.6  & -    & 11.0  & -  &   -   &    -  &  -    &   -   &   -   &  -\\
          & FCT~\cite{han2022fct} & Faster R-CNN & 17.1    & 21.4  & -  & -    & -  & -  &   -   &    -  &  -    &   -   &   -   &  -\\
          & QA-FewDet~\cite{hanQuery2021} & Faster R-CNN & 11.6  & 16.5  & 23.9  & 31.9  & 9.8   & 15.5  &   -   &   -   &    -  &  -    &   -   &  -\\
          & GCN-FSOD~\cite{han2022few} & Faster R-CNN & 12.8  & 15.2  & 24.0  & 27.7  & 12.3   & 14.7  &   2.7   &   3.7   &    13.8  &  16.0    &   22.4   &  25.6\\
          & DCNet~\cite{huDense2021} & Faster R-CNN & 12.8  & 18.6  & 23.4  & 32.6  & 11.2  & 17.5  & 4.3   & \textbf{\color{blue}6.9}   & 13.8  & 16.5  & 21.0    & 27.4 \\
          & Meta R-CNN~\cite{yanMeta2019} & Faster R-CNN & 8.7   & 12.4  & 19.1  & 25.3  & 6.6   & 10.8  & 2.3   & 2.8   & 7.7   & 11.6  & 14.0    & 19.0 \\
    \midrule
    \multicolumn{1}{c}{\multirow{5}{*}{\rotatebox{90}{\makecell{Algorithm-\\Oriented\\ Methods}}}} & TFA~\cite{wangFrustratingly2020}   & Faster R-CNN & 10.0    & 13.7  &-      &   -   & 9.3   & 13.4  &   -   &   -   &   -   &   -   &-       & - \\
        & Retentive R-CNN~\cite{fanGeneralized2021} & Faster R-CNN & 10.5  & 13.8  &    -  &   -   &   -   &   -   &   -   &   -   & -     &    -  &   -   &  -\\
        & MetaDet~\cite{wangMetaLearning2019} & Faster R-CNN & 7.1   & 11.3  & 14.6  & 21.7  & 6.1   & 8.1   & 1.0     & 1.1   & 4.1   & 6.2   & 12.2  & 17.3 \\
        & DeFRCN~\cite{qiaoDeFRCN2021} & Faster R-CNN & 18.5 & 22.6 &    -  &   -   &   -   &   -   &   -   &   -   & -     &    -  &   -   &  -\\
        & CFA~\cite{guirguis2022cfa} & Faster R-CNN & 18.9 & 22.6 &    -  &   -   &   -   &   -   &   -   &   -   & -     &    -  &   -   &  -\\
    \bottomrule
    \multicolumn{1}{c}{\multirow{3}{*}{}} & Faster R-CNN-ft*~\cite{yanMeta2019}   & Faster R-CNN & 5.5    & 7.4  &  10.0    &   13.1   & 5.5   & 7.4  &   -   &   -   &   -   &   -   &-       & - \\
          & Lee et al.*~\cite{leeFewShot2021} & Faster R-CNN & 13.4  & 17.1  &    30.6  &   35.2   &   9.1   &   14.7   &   -   &   -   & -     &    -  &   -   &  -\\
          & FCT*~\cite{han2022fct} & Faster R-CNN & 15.3  & 20.2  &    -  &   -  &   -  &   -   &   -   &   -   & -     &    -  &   -   &  -\\
          & TFA*~\cite{wangFrustratingly2020} & Faster R-CNN & 9.1  & 12.1  &    17.1  &   22.2   &   8.8   &   12.0   &   -   &   -   & -     &    -  &   -   &  -\\
          & Yang et al.*~\cite{yang2022efficient} & Faster R-CNN & 11.8  & 16.0  &    22.1  &   29.8   &   11.4   &   15.4   &   4.8   &   7.1   & 11.8     &    16.6  &   17.0   &  22.6\\
          & FSDetView*~\cite{xiaoFewShot2020} & Faster R-CNN & 12.5   & 14.7  & 27.3  & 30.6  & 9.8   & 12.2   & 2.5  & 3.2  & 13.8   & 15.2   & 19.9  & 23.8 \\
          & DeFRCN*~\cite{qiaoDeFRCN2021} & Faster R-CNN & 16.8 & 21.2 &    -  &   -   &   -   &   -   &   -   &   -   & -     &    -  &   -   &  -\\
    \bottomrule
    \end{tabular}}%
  \label{tab:coco}%
\end{table}

Table~\ref{tab:coco} provides more detailed information on FSOD performance under different scenarios.
Comparing AP$_{50}$ and AP$_{75}$, existing FSOD methods still suffer from the inaccurate location problem, as the performance drops significantly from AP$_{50}$ to AP$_{75}$.
Considering detecting objects of different scales, the performance on small objects (AP$_{S}$) is much worse than on large objects (AP$_{L}$).
For instance, taking the state-of-the-art method proposed by Park et al.~\cite{yanMeta2019} as an example, AP$_{L}$ of 10-shot detection is larger than AP$_{S}$ of 26.1 points, almost 4 times.
FSOD methods still need to consider coping with the localization and scale-variance problems, which are common in generic object detection.

As for the framework, from Table~\ref{tab:voc} and Table~\ref{tab:coco} we can see that most FSOD methods adopt Faster R-CNN~\cite{ren2015faster} as the base detector.
Faster R-CNN with the backbone of ResNet-50~\cite{he2016deep} and FPN~\cite{lin2017feature} achieves 33.9 AP and 56.9 AP50 on MS-COCO \textit{minival} set,
while the state-of-the-art method proposed by Park et al.~\cite{yanMeta2019} achieves 25.0 AP and 45.2 AP50 under the 30-shot setting.
Therefore, there is still a non-negligible gap between state-of-the-art FSOD models and generic object detection models trained with sufficient data.

As for training complexity, FSOD methods based on Faster R-CNN~\cite{ren2015faster} with FPN~\cite{lin2017feature} generally train the model on the base set for about 12 epochs.
Comparatively, FS-DETR~\cite{bulat2022fs} pre-trains the model on an extra dataset for 60 epochs and then trains the model on the base set for 30 epochs,
while Meta-DETR~\cite{zhangMetaDETR2021} takes 50 epochs to conduct the base training.
But they take similar time for training on both base and novel set afterward.
For parameters and floating-point operations per second (FLOPS), Faster R-CNN~\cite{ren2015faster} with ResNet-101 and FPN~\cite{lin2017feature} has 60M parameters and achieves 180 GFLOPS, while Deformable DETR~\cite{zhu2020deformable} has 40M parameters and achieves 173 GFLOPS.
Therefore, most FSOD methods based on these two frameworks have roughly similar model size and complexity.

\section{Technical Challenges}\label{sec:challenges}
Different from FSC, FSOD needs not only to recognize objects in the image but also to localize them.
Thus, it is impractical to directly adopt FSC methods to address the FSOD problem.
In FSC, each image can pass through a shared backbone, and then gets a fixed-length vector.
All subsequent operations can be done on this vector, which is very simple.
However, in FSOD an image may contain objects of different categories, and the feature of each object cannot be represented by the whole image but via the RoI features.
Currently, FSOD methods mainly adopt existing generic object detection models like Faster R-CNN~\cite{ren2015faster} and YOLO~\cite{redmon2016you} as a baseline, and then introduce or develop some FSL strategies.
In combining these two techniques, we still face some challenges:
\begin{itemize}
	\item \textbf{Overfitting on novel objects}.
	When generalizing the pre-trained detection model to novel classes, there is always a non-negligible performance drop on base classes.
	Moreover, due to lacking sufficient training samples, the model is prone to overfit on novel classes.
	The support set contains limited information, thus it is challenging to learn the original novel class distribution.
	Some categories are highly intra-diverse, which makes FSOD more difficult.
	\item \textbf{Scale variation}.
	Scale variation is a common challenge in object detection~\cite{lin2017feature, ghiasi2019fpn}, but is more serious in FSOD owing to the unique sample distribution~\cite{wuMultiscale2020}.
	FPN~\cite{lin2017feature} is widely adopted in existing generic object detection methods and FSOD methods.
	Nevertheless, it aims at handling objects of different scales at different levels, rather than enriching the scale space.
	Desirable FSOD methods should be capable of detecting objects of various scales with a narrow scale space.
	\item \textbf{Class confusion}.
	Object detection is generally decomposed into classification and regression tasks, where the former is category-specific while the latter is category-agnostic.
	It was shown in some FSOD works~\cite{sunFSCE2021, liFewShot2021a} that the main source of false positives is misclassifying objects into similar categories instead of inaccurately localizing them.
	In the real world, many objects are similar in appearance that even humans can get confused.
	Without sufficient training samples, it is difficult for the model to generate discriminative and generalized representations for each class, which leads to poor classification results on confusable classes.
	\item \textbf{Robustness to cross-domain detection}.
	FSOD models are firstly trained on the \textit{base set}, which brings the domain shift problem, i.e., the \textit{base set} may share little knowledge with the \textit{novel set}.
	Under such circumstance, the widely-adopted anchor-based models cannot generate favorable proposals of novel classes, leading to performance degradation~\cite{kangFewShot2019}.
	Wang et al.~\cite{wangFewShot2019} adapted Faster R-CNN~\cite{ren2015faster} to tackle the cross-domain detection problem, but the source domain may share similar categories with the target domain. Domain shift in FSOD remains an open problem to be solved.
\end{itemize}

\section{Future Directions}\label{sec:directions}
FSOD is an emerging research topic in computer vision.
There is still a non-negligible performance gap between FSOD and generic object detection that utilizes abundant data to train the model.
Here, we highlight some future research directions of this area.
\begin{itemize}
	\item \textbf{FSOD-specific frameworks}.
	Existing FSOD methods adopt a generic object detection framework like Faster R-CNN~\cite{ren2015faster} as the base detector, which is not designed specifically for the few-shot scenario and thus limits the detection performance.
	It is significant to design frameworks suitable for the few-shot problem.
	Meta-learning is popular in tackling the few-shot classification problem, and some FSOD methods~\cite{kangFewShot2019, zhangAccurate2021, zhangMetaDETR2021} introduce it to FSOD.
	However, most meta-learning-based FSOD methods are memory-inefficient, i.e., can only detect one class in the query image at one go.
	Therefore, novel meta-learning frameworks for FSOD are still needed.
	\item \textbf{More robust detectors}.
	As shown in Tab.~\ref{tab:voc}, most existing FSOD methods suffer from significant performance drop when the number of shots decreases.
	Moreover, Fig.~\ref{fig:bar_voc} and Fig.~\ref{fig:bar_coco} show that the performance may vary largely for different training samples.
	To achieve more stable performance over different training datasets, robust feature extractors that can generate more general features for each class are desirable.
	\item \textbf{More accurate localization}.
	Few-shot learning for image classification has achieved momentous progress in recent years.
	Many FSOD methods convert FSOD into an FSC task while paying not too much attention to the localization.
	However, Tab.~\ref{tab:coco} shows that there is a huge gap between AP$_{50}$ and AP$_{75}$ for all the compared methods.
	Thus, it is insufficient to merely focus on classification while ignoring regression.
	\item \textbf{Multi-scale FSOD}.
	Scale variation is a common problem in computer vision tasks, and many generic object detection approaches dedicate to solving this problem~\cite{he2015spatial, lin2017feature, tan2020efficientdet, liu2018path, ghiasi2019fpn, singh2018sniper}.
	It is shown in Tab.~\ref{tab:coco} that the performance on small (AP$_{S}$), medium (AP$_{M}$), and large (AP$_{L}$) objects varies a lot.
	As aforementioned, the scale variation problem is more challenging in FSOD.
	Novel FSOD methods should consider this problem more.
	\item \textbf{Data augmentation with corpus}.
	Each class can be explicitly described by the text.
	In another word, semantic description is invariant while visual representation is variant for the same class.
	Some methods~\cite{zhuSemantic2021, cao2021few} demonstrate the potential of utilizing semantic information obtained from large corpora is helpful to improve FSOD performance.
	They focused on class relations, but other relations such as scene-object relations may also be beneficial to both classification and localization.
	\item \textbf{Open-world detection}.
	The ultimate goal of object detection is to develop a model capable of detecting almost all categories in the open world and learning fast from unseen classes.
	Joseph et al.~\cite{joseph2021towards} devised a novel object detection framework to achieve this purpose, but did not consider the classes with few samples.
	P$\textnormal{\'e}$rez-$\textnormal{\'R}$ua et al.~\cite{perez-ruaIncremental2020} and Li et al.~\cite{liGeneralized2021} presented iFSOD models, partially tackling the open-world detection problem.
	So there is still an urgent need for more general and incremental methods of open-world detection.
\end{itemize}

\section{Conclusion}\label{sec:conclusion}
Few-shot object detection (FSOD), as an application of few-shot learning (FSL), intends to achieve competitive performance with the human visual system when target classes contain only a few samples.
In this paper, we provide a comprehensive review of the recent achievements of FSOD.
First, we give a brief introduction to FSL and generic object detection, including their problem definition and recent advances.
Then, we present an overview of FSOD from the perspectives of problem definition, common datasets, evaluation metrics, and a new taxonomy of FSOD methods, under which the
existing FSOD works are grouped into three types: data-, model-, and algorithm-oriented methods in terms of what major contribution each method makes in three aspects: data, model, or algorithm.
Following that, major advances in FSOD are reviewed, and their performances are analyzed and compared.
Finally, we highlight the main technical challenges and future research directions. Considering that FSOD is still an emerging and promising research computer vision topic, this survey may serve as a start point for researchers who are interested in this topic.

\bibliographystyle{ACM-Reference-Format}
\bibliography{ref}


\end{document}